\documentclass[sigconf]{acmart}
\AtBeginDocument{%
  \providecommand\BibTeX{{%
    \normalfont B\kern-0.5em{\scshape i\kern-0.25em b}\kern-0.8em\TeX}}}

\usepackage[some,bottom]{background}
\usepackage{stfloats}

\copyrightyear{2024}
\acmYear{2024}
\setcopyright{rightsretained}
\acmConference[DIS '24]{Designing Interactive Systems Conference}{July 1--5, 2024}{IT University of Copenhagen, Denmark}
\acmBooktitle{Designing Interactive Systems Conference (DIS '24), July 1--5, 2024, IT University of Copenhagen, Denmark}\acmDOI{10.1145/3643834.3660721}
\acmISBN{979-8-4007-0583-0/24/07}

\acmSubmissionID{5209}

\begin{document}

\newcommand{\eg}[0]{\textit{e.g.,}}
\newcommand{\ie}[0]{\textit{i.e.,}}

\newcommand{\inlinequote}[1]{{``\emph{#1}''}}

\newcommand{\modified}[1]{{\color{black} #1}}
\newcommand{\finalfix}[1]{{\color{black} #1}}

\title{Understanding On-the-Fly End-User Robot Programming}

\author{Laura Stegner}
\authornote{Both authors contributed equally to this research.}
\orcid{0000-0003-4496-0727}
\affiliation{%
  %%%%\institution{Computer Sciences Department}
  \institution{University of Wisconsin–Madison}
  \city{Madison}
  \state{Wisconsin}
  \country{United States}
  \postcode{53706}
}
\email{stegner@cs.wisc.edu}

\author{Yuna Hwang}
\authornotemark[1]
\orcid{0000-0001-7726-8003}
\affiliation{%
  %%%%\institution{Computer Sciences Department}
  \institution{University of Wisconsin–Madison}
  \city{Madison}
  \state{Wisconsin}
  \country{United States}
  \postcode{53706}
}
\email{yunahwang@cs.wisc.edu}

\author{David Porfirio}
\orcid{0000-0001-5383-3266}
\affiliation{%
  %%%%\institution{Navy Center for Applied Research in Artificial Intelligence}
  %%%%\institution{NRC Postdoctoral Research Associate}
  %\institution{NRC Postdoctoral Research Associate}
  \institution{U.S. Naval Research Laboratory
  % \\NRC Postdoctoral Research Associate
  }
  \city{Washington}
  \state{DC}
  \country{United States}
  \postcode{20375}
}
\email{david.j.porfirio2.ctr@us.navy.mil}

\author{Bilge Mutlu}
\orcid{0000-0002-9456-1495}
\affiliation{%
  %%%%\institution{Computer Sciences Department}
  \institution{University of Wisconsin–Madison}
  \city{Madison}
  \state{Wisconsin}
  \country{United States}
  \postcode{53706}
}
\email{bilge@cs.wisc.edu}


\begin{abstract}
%EUD is in the "grab bag" of techniques
%EUD is "necessary" --> e.g., caregivers want oversight over robot, enforce boundaries/check what robot 
%Highlight that Tabula uses rudimentary AI planning, but still need goal from the user/any constraints/preferences
%Missing: make the case for EUD

  Novel end-user programming (EUP) tools enable on-the-fly (i.e., spontaneous, easy, and rapid) creation of interactions with robotic systems. These tools are expected to empower users in determining system behavior, although very little is understood about how end users perceive, experience, and use these systems. In this paper, we seek to address this gap by investigating end-user experience with on-the-fly robot EUP. We trained 21 end users to use an existing on-the-fly EUP tool, asked them to create robot interactions for four scenarios, and assessed their overall experience. Our findings provide insight into how these systems should be designed to better support end-user experience with on-the-fly EUP, focusing on user interaction with an automatic program synthesizer that resolves imprecise user input, the use of multimodal inputs to express user intent, and the general process of programming a robot.
    
  %To enable the easy deployment of robots that aid people with daily tasks, robot end users will require tools that facilitate on-the-fly (i.e., spontaneous, easy, and rapid) task creation. Although recent research has introduced highly capable end-user programming (EUP) tools, there is a lack of understanding of how end users perceive, experience, and use them. We thereby seek to help close this gap by investigating end users' experiences with on-the-fly robot EUP. In this paper, we trained 21 end users to use an existing on-the-fly EUP tool, asked them to create robot programs for four scenarios, and assessed their overall experience. Our findings provide insight into end users' experiences with different facets of on-the-fly EUP, such as interacting with a program synthesizer that automatically completes missing elements of programs, the use of multimodal inputs to express their intent, and their general approach to the robot programming process.
\end{abstract}



% \ccsdesc[300]{Computer systems organization~Robotics}




\maketitle

\section{Introduction}

\begin{figure}[!t]
  \includegraphics[width=\columnwidth]{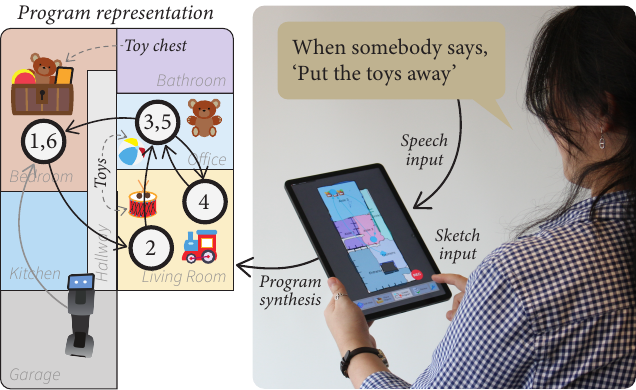}
  \caption{We investigate end-user experience with on-the-fly robot end-user programming using \textit{Tabula}, a state-of-the-art open-source research prototype. \textit{Right:} An experimenter using speech and touch input to program a robot to put toys away in a toy chest. \textit{Left:} A visual representation of the generated program by a study participant (P5).}
  \Description{A woman is holding a tablet in one hand. The tablet shows a map floorplan. A speech bubble coming from the woman says "when somebody says, 'Put the toys away.'" An arrow labeled "speech input" points to the tablet. The tablet is labeled "sketch input." An arrow labeled "program synthesis" goes to an enlarged view of the map floorplan. The floorplan is labeled "program representation" and contains numbered dots in the following labeled regions of the floorplan: 1,6 bedroom; 2 living room; 3,5 office; 4 living room. The floorplan also contains the following labeled objects: toy chest in bedroom, toys in office and living room.}
  %On the left, a caregiver is holding a tablet and standing facing a robot. The label says "programming interface." On the right, two scenes in a senior living community are shown. The first is labeled "scheduled care robot task" and shows a robot delivering a coffee cup to an elderly woman sitting in a chair. The second is labeled "More time for caregiver" and shows a caregiver smiling at an elderly man.
  \label{fig:teaser}
\end{figure}

% motivate need for EUP
Robots are increasingly being designed to aid \emph{end users} in completing day-to-day tasks. These robots arrive with autonomous capabilities, yet they still require input from end users about which tasks must be completed and any contextual details surrounding the task. End users could include residents with robots in their private homes \cite{lu2012structure}, shopkeepers with robot assistants to aid customers \cite{senft2020would}, caregivers with robots to assist in providing care to residents \cite{stegner2023situated}, and many more examples. In each of these scenarios, the end user may need to communicate to the robot a task for it to complete \textit{on the fly}, \ie{} spontaneously, easily, and rapidly. To address this need, researchers have created various \emph{end-user programming} (EUP) tools to allow end users to create interactions with robotic systems without extensive technical knowledge \cite{ajaykumar2021survey}. Specifically, EUP tools produce robot programs, which traditionally consist of sequences of actions for the robot to perform in order to complete a task. 

Methods, techniques, and tools that facilitate rapid and intuitive robot EUP are rapidly proliferating (see \citet{ajaykumar2021survey} for a detailed review of robot end-user programming), including tools that better capture user intent \citep[\eg{}][]{cao2019v}, automatically synthesize programs given high-level user input \cite{porfirio2019bodystorming}, or contextualize programs within the user's environment \cite{huang2020vipo}. EUP tools that incorporate multimodal inputs \citep[\eg{}][]{forbes2015robot, stenmark2017simplified,porfirio2021figaro,porfirio2023sketching} often combine various methods and techniques in an effort to create a more intuitive and natural on-the-fly EUP experience.

Despite recent advances in EUP tools for robot programming, their full potential and impact remain unknown. A rich understanding of user experience with state-of-the-art EUP tools is missing, as these advanced EUP systems have yet to find real-world use and the research literature lacks \finalfix{deep} understanding of use patterns, user experience, and limitations of these systems. 
As a result, very little is known about how these tools might be used by end users. Increasingly sophisticated methods and techniques that deviate from traditional programming paradigms require further exploratory user studies to contextualize the technical advances within end user needs and experiences.
 
Therefore, to help close this gap, we conducted an in-depth exploratory evaluation using a state-of-the-art on-the-fly EUP prototype called \emph{Tabula} \cite{porfirio2023sketching}. Tabula is an open-source EUP prototype tool \modified{that we developed previously (see \citet{porfirio2023sketching}). It} facilitates on-the-fly robot programming \finalfix{by combining }multimodal input \finalfix{that enables} end users to express task intent \finalfix{with a} program synthesis \finalfix{technique} that \finalfix{automatically }completes missing elements of a program. Using Tabula as a medium for creating robot programs, we investigate the following research question.

\begin{itemize}
    \item \textbf{RQ:} What are end users' experiences with on-the-fly robot programming?
\end{itemize}

To answer this question, we trained 21 participants to use Tabula and instructed each participant to create robot programs for three structured robot scenarios and one open-ended robot scenario. 
Specifically, we consider how end users approach the on-the-fly robot programming process through an in-depth exploratory evaluation of Tabula's key features, the multimodal inputs, and the program synthesizer. Therefore, we have the opportunity to probe usability aspects specific to Tabula's implementation, but also glean more widely applicable design insights.

This paper contributes to understanding how end users approach robot EUP through (1) a user study that evaluates a multimodal, on-the-fly EUP tool; (2) five themes that relate to both the usability and design of on-the-fly robot programming tools; and (3) a set of design guidelines that can inform future on-the-fly EUP tool design.

\section{Related Work}
Our work builds on prior literature from software engineering, human-computer interaction (HCI), and human-robot interaction (HRI), focusing on how end users specify requests to interactive systems, approaches to end-user development and programming, and prior studies of end-user programming (EUP) tools.
% to specify system behavior, as summarized below. 

\subsection{Approaches to End-user Specification}
Many programs written today do not rely entirely on professional programmers or roboticists \cite[\eg{}][]{de2015homerules, ko2011state, wong2007making, maclaurin2009kodu, gorostiza2011end}. Instead, end users with discrete domain expertise drive software development, specifically by contributing to obtaining a complete and consistent set of system requirements \citep[]{kassel2003approach, van1998experiences}. Thus, seminal work in the software engineering field \cite[\eg{}][]{barricelli2019end, ko2011state} provides pointers \finalfix{on} how to facilitate end-user specification, particularly at the exploratory phase \cite{gomaa1981prototyping} of the software lifecycle. \emph{Dialogue} is an accessible paradigm for rapid prototyping based on its use in daily human communication \cite{ajaykumar2021survey}. \citet{porfirio2019bodystorming} proposed an approach that utilized speech gathered from ``role-playing'' to synthesize human-robot interaction scenarios. Within the end-user specification frame, \emph{visual programming interfaces} are frequently utilized. Flow-based visual interfaces allow users to conceptualize programs as processes \cite{zarrin2015towards}. In RoboFlow \cite{alexandrova2015roboflow}, edits to default programs can be easily made with the assistance of a flow-based visual expression. 

Display of readily distinguishable domain-specific operation units to end users has proven successful when deployed on a visual interface. The system implemented and evaluated by \citet{senft2021task} only exposes the graphical representation of the \emph{task-level} (high-level) actions to the user, which in turn allowed effective teleoperation of users for individuals with varying levels of expertise. More recently, deep learning and large-language modeling (\emph{LLM}) methods are gaining attention for ``prompt-based prototyping'' \citep[\eg{}][]{arawjo2023chainforge, jiang2022promptmaker}. ChatGPT (GPT-3.5 and GPT-4 \cite{openai2023gpt4}) and its related work \cite[\eg{}][]{ostyakova2023linguistic, benzon2023discursive} serve as distinct use cases where the representation format of  question-answer pairs closely resemble that of interpersonal communication, borrowing dynamics of turn-taking. 

%Another approach concerns the representation format that bridges the end users and the system, where domain-specific languages and graphical languages facilitate end-user specification \cite{barricelli2019end}. With \emph{dialogue} as an interaction format that faces the end user directly, the content of the dialogue can be parsed into clauses and subsequently into commands which can then be fed into robotic systems \cite{porfirio2023sketching}. 
%Display of readily distinguishable, domain-specific operation units to end users has been proven successful when deployed on a \emph{visual interface}. The system implemented and evaluated by \citet{senft2021task}, only exposes the graphical representation of the \emph{task-level} (high-level) actions to the user, which in turn allows effective teleoperation of users for individuals with varying levels of expertise. ChatGPT (GPT-3.5 and GPT-4 \cite{openai2023gpt4}) and its related work serve \cite[\eg{}][]{ostyakova2023linguistic, benzon2023discursive} as a distinct use case where the representation format of a question-answer pair closely resembles that of interpersonal communication, borrowing dynamics of turn-taking. 

\subsection{End-User Development and Programming}

\textit{End-user development} (EUD) encompasses tools and techniques that facilitate the creation of software systems by non-programmers \cite{lieberman2006end}.
Crucially, \citet{lieberman2006end} distinguished ``design-before-use'' EUD as creating software artifacts prior to their execution versus ``design-during-use'' as modifying existing software already in use.
\textit{End-user programming} (EUP) is a type of EUD that typically occurs at the creation phase.
Although both paradigms play important roles within robotics, the focus of programming tools for human-robot interaction is often on EUP, with these tools having distinct \textit{authoring} phases involving the initial creation of a program \cite{ajaykumar2021survey}.

EUP tools capture user intent in a variety of different ways, often taking the form of traditional keyboard-and-mouse visual programming environments \citep[\eg{}][]{schoen2022coframe, alexandrova2015roboflow, leonardi2019trigger}, demonstration \citep[\eg{}][]{huang2017code3, gao2019pati}, and, more recently, \textit{in situ} interfaces via mixed and augmented reality \citep[\eg{}][]{cao2019ghostar, cao2019v}.
Often, these interfaces require multimodal input from developers, such as \textit{Figaro} \cite{porfirio2021figaro}, in which users paired spoken language statements with physical demonstrations \finalfix{through} figurines.
Due to the nature of programming, however, EUP systems often require meticulous and clear input from the user, which can be awkward for users of multimodal systems \cite{porfirio2019bodystorming}.

The focus of our work is to better understand how end users naturally approach programming using EUP tools. % on how multimodal EUP tools in HRI should expect to capture \textit{natural} input from users.
Natural input is often imprecise and rapid, a key observation of the \textit{sloppy programming} paradigm \cite{little2010sloppy}.
Specifically, we focus on how end-user programmers might combine two historically popular EUP input modalities---\textit{spoken language} and \textit{sketching}.
Spoken language has experienced widespread popularity for programming HRI systems in the collaborative \cite{forbes2015robot} and service \cite{walker2019neural} domains.
Sketching, too, has seen success within HRI EUP \cite{liu2011roboshop, sakamoto2009sketch}, and has occasionally been paired with speech for robot control \cite{correa2010multimodal, teller2010voice}. Therefore, exploring how end users interact with these modalities within a working prototype will aid in the design of future EUD and EUP systems.
%\subsection{EUD evaluations}

% section about how people used these tools
\subsection{EUP Tool Usage}%in HCI/HRI}
A critical aspect of EUP research in HRI and HCI is investigating how EUP tools could be used.
Formative design studies are common practice in EUP to investigate the potential use of tools that have not been built yet.
Related to our work, \citet{li2019pumice} investigated how a touchscreen interface can enhance spoken language, and \finalfix{found} that multimodal input can reduce unclear or vague concepts in speech.
Other work used formative studies to investigate the potential applications for which a hypothetical EUP tool might be used \cite{chung2016iterative}.
In addition to formative studies, \citet{alves2022flex} presented a myriad of case studies documenting how their EUP tool was used in real-world and open-ended deployments, including how end users applied the tool for robot personalization.
Within the realm of general programming, \citet{puig2018virtualhome} provided information on the kinds of programs users could create for robots to perform when provided with an open-ended development environment.

In our review of related EUP literature, we note that most work highlights technical over empirical contributions.
%most evaluations are technical rather than empirical.
Technical contributions are often still accompanied by usability measures \modified{\citep[\eg{}][]{lucci2014understanding, buchina2019natural}} and measures of whether study participants are able to meet predetermined task criteria successfully \citep[\eg{}][]{huang2017code3}. \modified{Most work that makes empirical contributions performs summative evaluations, including either quantitative scales, such as the System Usability Scale (SUS), or the Cognitive Dimensions of Notations (CDN) \citep[\eg{}][]{buchina2016design,buchina2019natural}, or open-ended, qualitative findings \citep[\eg{}][]{porfirio2021figaro}. However, these empirical findings are often in service of validating the technical contributions of the work. }
%However, focusing on quantitative metrics neglects participant experiences that naturally emerge during the interaction with the EUP tool and confines the discussion of the EUP tool to a pre-determined set of usability measures \citep[\eg{}][]{buchina2016design, buchina2019natural}. To our knowledge, few open-ended explorations exist that emphasize a qualitative investigation of how participants use EUP tools.}  
%but few open-ended explorations of how participants use EUP tools.
\modified{In this work, we aim to add to the body of empirically focused EUP literature with a deeper understanding of user experience and use patterns with EUP tools and design guidelines derived from this understanding.}

\section{Method}
We conducted a user study where we asked participants to use a multimodal EUP research prototype, \textit{Tabula}.\footnote{All study materials, de-identified data, codebook, and supplementary video are available through the following OSF repository: \url{https://osf.io/ps2fw/}}

\subsection{Participants}
We recruited 21 individuals to participate in the study, aged 18--72 years ($M=25.19$ years, $SD=11.53$ years; 11 males, 10 females). While prior programming experience or exposure to robotic systems was not required, 12 participants reported previous programming experience ($M=3.92$ years, $SD=2.94$ years), and five of those participants also reported exposure to robotic systems ranging from using Lego robotics kits as a child to attending a Human-Computer Interaction summer school that included a robotics project. Participant backgrounds included 15 occupations or student majors from a variety of different fields which spanned science, engineering, math and statistics, medicine, and humanities.

\begin{figure}[!tb]
    \centering
    \includegraphics[width=\columnwidth]{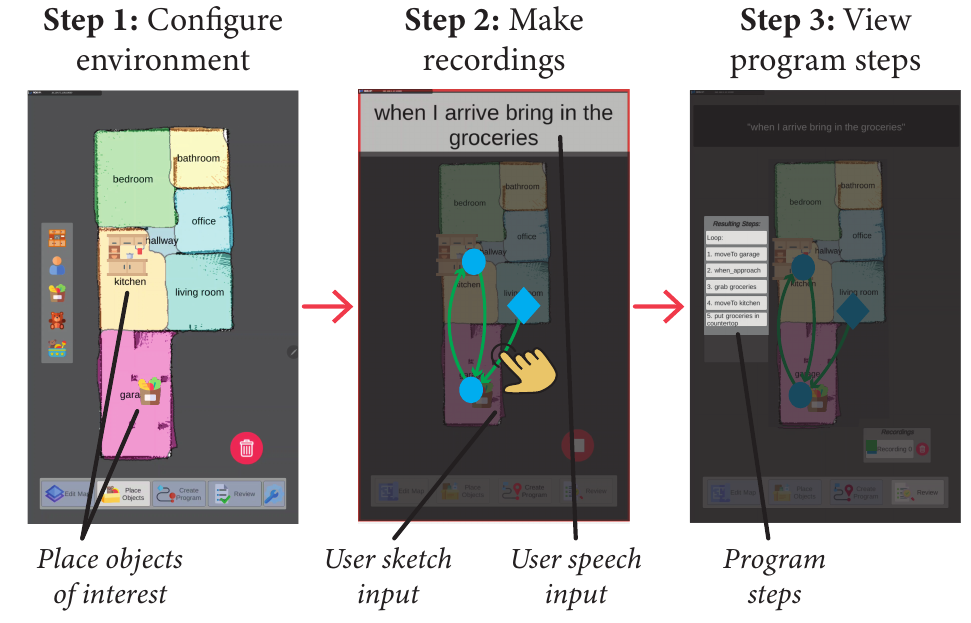}
    \caption{We used \textit{Tabula}, a multimodal EUP tool that uses a combination of speech and sketching input to generate a robot program \cite{porfirio2023sketching}, to study end-user experiences with programming robots on the fly. \textit{Left:} First, users configure the environment, including placing any objects for the robot to interact with. \textit{Middle:} Second, users create recordings by first providing a speech utterance to instruct the robot what to do and subsequently creating a sketch by drawing a path of points of interest that the robot should visit. \textit{Right:} Finally, inputs are combined by the program synthesizer, and users can view the resulting programming steps.}
    \Description{Three horizontal panels, each linked from left to right by an arrow. The first panel is labeled 'Step 1: Configure environment' and contains a screenshot of the Tabula interface shows a map floorplan of an apartment which includes 6 rooms: garage, kitchen, living room, hallway, office, bedroom, bathroom. There is a cabinet in the kitchen and a bag of groceries in the garage. Both are labeled 'Place objects of interest.' The second panel is labeled 'Step 2: make recordings' and shows the same map floorplan, but the top of screen shows text that says 'when I arrive bring in the groceries' and labeled 'User speech input,' and the floorplan contains a touch icon overlaid on a set of dots connected by arrows in different regions of the map . The touch icon is labeled 'user sketch input.' The third panel is labeled 'Step 3: View program steps' and contains the same image as panel 2 but now overlaid with vertically stacked boxes in the left. The boxes contain illegible text and are labeled 'Program steps.'}
    \label{fig:system}
    \vspace{-9pt}
\end{figure}

\subsection{Interface}
For our study, we \modified{used} an open-source, state-of-the-art research prototype tool called \emph{Tabula} \modified{which we developed in previous work (see \citet{porfirio2023sketching})}. 
Tabula is a handheld EUP tool where given a 2-dimensional bird's-eye view of an environment, users can utilize multimodal speech and touch inputs to create custom robot programs \cite{porfirio2023sketching}. The user can first optionally configure the environment by placing relevant objects (\eg{} toys, cabinets) with which the robot could interact. Then, the user creates the robot program by creating one or more \emph{recordings}, which consist of a combination of a spoken command and a sketched path drawn on the interface. Their utterance is parsed into the core of what the robot has to achieve, including the base command (\eg{} put, move to) and any relevant parameters for that command (\eg {} objects or places within the environment). The drawn sketch includes a series of waypoints that  represents locations the robot must visit during program execution. The system then contextualizes the \emph{core} (the user's command and its parameters) within the drawn path\finalfix{, culminating in} a program \finalfix{with} waypoints \finalfix{from} the \finalfix{sketch}.
%embedding program logic onto the two-dimensional environment representation. 
Users are able to view the final program steps after creating recordings through a separate review panel. A high-level system operation is presented in Figure \ref{fig:system}.

In the version of Tabula that we used, users do not meticulously specify \textit{step-by-step} programs (\ie{} they do not specify commands and locations in the exact order to be performed), but rather supply the system with the core of the program and then contextualize that core with the sketch.\footnote{Tabula's implementation does not restrict users to providing the speech core and sketched path in any particular order, but its compilation to Android for this study imposes this restriction.} The command extracted from the utterance is not guaranteed to happen at any specific location, as the synthesizer will automatically decide where to place commands within the sketch. Therefore, we describe this input as \textit{non-sequential}.

For example, the user may utter the speech ``put the groceries in the kitchen'' and draw a path to the garage, then to the kitchen. The system inferred that in order to `put' the groceries, it first needs to `grab' them. Since there are two actions, the system also infers that it should `grab' at the first location and `put' at the second. The system further has the constraint that the `put' command requires a container to put the object in\finalfix{---}in this case, putting the groceries in the kitchen cabinets. While the utterance did not include the container parameter, the system infers which container based on a pre-configured dictionary of what containers are in different locations. The system would therefore interpret the given inputs such that the robot should travel to the garage, grab the groceries, travel to the kitchen, and put the groceries in the kitchen cabinet.

During a basic operation of Tabula, users create one recording which results in a basic robot program that accomplishes at most one goal based on the core. However, by creating multiple recordings, Tabula enables end users to specify more complex logic, \ie{} branching and looping. To create a branch, the user creates a second recording that starts from an existing waypoint and includes a \textit{trigger} speech (\eg{} ``when I arrive...'') to indicate when the robot should opt to follow that branch. To create a loop, the user simply returns to a previously-visited waypoint within one recording. Specifying a loop's exit condition requires a second recording where the trigger speech indicates the desired exit condition, \eg{} ``when I say stop...''

Tabula was selected due to its inclusion of state-of-the-art research concepts described above that have not yet been widely evaluated by end users. The key features include: (1) the combination of speech and touch input, (2) the automatic completion of an under-specified user input by searching for adequate entities to satisfy relevant preconditions, and (3) the embedding of programming logic (\eg{} loops) within the program to address task complexity. \citet{porfirio2023sketching} specify that these features are intended to remove some of the users' burden in constructing comprehensive, end-to-end robot programs. However, as \citet{porfirio2023sketching} do not include a user study that examines the usability of the system, our evaluation aims to better understand precisely how these features support end-user programming efforts. 

\subsection{Procedure}
Participants were guided to a quiet room for the study. The experimenter briefly introduced what the study would entail and then the participants provided their informed consent before continuing. This study was reviewed and approved by the \modified{University of Wisconsin--Madison} Institutional Review Board (IRB). The study consisted of the following five phases:

\paragraph{Tutorial \& Training}
Participants learned Tabula through 26 minutes of tutorial videos designed to help participants to familiarize themselves with the basic operations of the system. During the tutorial, the interface used a supermarket environment. The tutorial session was interactive, meaning the experimenter paused the video at pre-set points to prompt participants to try the examples from the tutorial. For example, participants practiced making recordings with the \textit{speech} ``say hello follow me to the sale'' and a \emph{sketch} of a path from anywhere on the map to the entrance of the store. The tutorial videos also asked participants questions to check their understanding of the key system rules. For example, the tutorial was designed deliberately to build upon previous concepts and raised questions on the difference between the new constraint and the previous constraint (\eg{} how does adding the new speech ``if someone says yes'' change how you program the robot?). Overall, the tutorial delivered the logistics of how to make a command with respect to the interface (\eg{} using which modality to specify a complete command) and provided examples of use cases where the system supports programming logic (\eg{} loops).

\paragraph{Structured Scenarios}
Participants were then prompted to work with four different design scenarios. Participants programmed a human-robot interaction using the interface. We then asked participants to \emph{think aloud} while completing each scenario, which allowed the experimenter to notice any hesitancy from the participant and ask clarifying questions thereafter. All three structured scenarios commonly used a home environment. We deliberately used a different environment in the scenarios versus the tutorial to observe how participants were to operate with the system, without relying on their familiarity with a specific environment. 
The scenarios were designed around the idea that participants were hosting a party and wanted the robot to help prepare. For each scenario, participants were briefed on the context and given an objective of what the robot program should accomplish. The objectives encouraged participants to use a variety of Tabula's features, including a robot passing an object, carrying multiple objects, and acting in regard to varying responses from the end-users described in the scenario. The comprehensive list of objectives that the participants were asked to complete is as follows.
\begin{itemize}
    \item \emph{Scenario 1}: The robot should put away the toy
    \item \emph{Scenario 2}: The robot should bring all of the groceries from the garage to the kitchen
    \item \emph{Scenario 3}: The robot should respond to guests that approach it to either show them the kitchen or the bathroom
    %\item[\emph{Scenario 1}] You are out at the grocery store, and you realize your dog left its toy teddy bear lying around in the bedroom, and you want the robot to put it away in the living room.
    %\begin{itemize}
    %    \item \emph{Objective}: The robot should put away the toy
    %\end{itemize}
    % \item[\emph{Scenario 2}] You had a busy morning at the store, buying lots of food for the event! You want the robot to help unload the groceries. They will be in the garage, and you want the robot to bring all of the groceries to the kitchen. Since it’s for a party, you got a lot of food, probably more than anyone can carry at once.
    %\begin{itemize}
    %    \item \emph{Objective}: The robot should bring all of the groceries from the garage to the kitchen
    %\end{itemize}
    %\item[\emph{Scenario 3}] You realize during the party, you want the robot to be able to help guests with directions. The main party is in the living room, but people may need to go to either the kitchen or the bathroom. If a person asks where is the bathroom, what should the robot do? If the person asks where is the kitchen, what should the robot do? In this case, the robot cannot point because it is too crowded, so it must physically move to the location to help with directions. But, we don’t need to worry whether the person is following or not.
    %\begin{itemize}
    %    \item \emph{Objective}: The robot should respond to guests that approach it to either show them the kitchen or the bathroom.
    %\end{itemize}
\end{itemize}

\paragraph{Open-Ended Scenario}
After the participants completed the three structured scenarios, the experimenter then asked the participants to come up with their own scenarios. Because these scenarios were open-ended, participants were given a choice to use either the supermarket environment or the home environment. 

\paragraph{Interview \& Questionnaires}
Following the open-ended scenario, the experimenter asked participants to respond to a usability questionnaire (the System Usability Scale (SUS) \cite{brooke1996sus}) based on their experience across all scenarios.
In the last portion of the study, the experimenter conducted a semi-structured interview and asked participants to respond to a demographic questionnaire. The interview questions include topics such as the perceived level of system flexibility (\eg{} if the participants deemed the rules of the system too rigid) and the dynamic participants experienced while utilizing both speech and touch (\eg{} if the order of operation of the speech first and sketch second was natural for them). 

\subsection{Measures and Analysis}
We collected the following data: 10 items from the SUS questionnaire \cite{brooke1996sus}, screen recordings of tablet usage during the scenarios, audio recordings of the think aloud conducted during the scenarios, and audio recordings of responses to questions during the semi-structured interviews. The think aloud and interviews were transcribed and formatted into tables for analysis. Two coders reviewed the data and decided to split the analysis of the think alouds and interviews due to the additional context required to understand the think alouds, as that dialogue is tightly linked to participants' use of the interface. For the interview transcripts, one coder developed a codebook and conducted a thematic analysis following the guidelines of \citet{braun2006using}. For the scenario data, one coder developed a codebook for the think aloud transcripts and a list of behaviors to code in the screen recordings, \eg{} started a recording, made a speech input, checked the review mode, \textit{etc}. Screen recording data was coded using BORIS \cite{friard2016boris}, an open-source event recording software. The think aloud and screen recording data was then chronologically organized and subsequently analyzed for frequency of co-occurring codes within a five-second window. Across all coding, the two coders had a high inter-rater reliability (Cohen's Kappa, $\kappa=0.83$), which indicates an ``almost perfect'' agreement according to interpretation guidelines from \citet{landis1977measurement}. We present themes that emerged through the interview data codebook as well as the patterns that emerged from the scenario data.

\begin{table*}[!t]
\caption{A summary of the themes developed in our analysis.}
\label{tab:theme_summary}
\centering\small\renewcommand{\arraystretch}{1.1}
\begin{tabular}{ p{0.98\textwidth} } 
    \toprule
    \textbf{Summary of Findings} \\ 
    \midrule
    \addlinespace[.25cm]
    \textit{General} --- These themes relate to user experiences which may generalize to other on-the-fly EUP tools. \\
    \addlinespace[.15cm]
    \textbf{Theme 1: End users viewed program steps to better understand the system} \\ End users relied heavily on viewing program steps to shape their understanding of how the system works and to look ahead to their next actions. \\
    \addlinespace[.15cm]
    \textbf{Theme 2: End users have poor mental model of the input paradigm} \\ End users who naturally tended toward step-by-step instructions for the robot struggled with articulating their intent non-sequentially. \\
    \addlinespace[.15cm]
    \textbf{Theme 3: End users felt that the robot was a tool to use} \\ End users viewed creating the robot programs as a way to utilize a robot in a tool-like manner rather than as an independent, autonomous agent. \\
    \midrule
    \addlinespace[.25cm]
    \textit{Usability} --- These themes relate to specific experiences based on Tabula's implementation of on-the-fly EUP. \\
    \addlinespace[.15cm]
    \textbf{Theme 4: End users had mixed experiences on interaction with the program synthesizer} \\ End users either appreciated that the program synthesizer provided human-like common sense support, or they disliked the assistance because they perceived it as a loss of control over the robot program. \\
    \addlinespace[.15cm]
    \textbf{Theme 5: There is more to using the system than understanding its basic functionality} \\ Even after learning Tabula's basic functionality, end users still faced a learning curve to become proficient with its use. \\
    \bottomrule
\end{tabular}
\end{table*}

\section{Findings}\label{sec:findings}
Participants overall had a positive experience with the interface, with a ``good'' \cite{bangor2009determining} mean SUS score of $69.9$ ($Median=72.5$, $SD=11.6$). 
From our qualitative analysis, we developed five themes about how end users perceived and interacted with various features of the on-the-fly end-user programming (EUP) tool that they used. The themes are summarized in Table~\ref{tab:theme_summary}. Themes 1--3 illustrate experiences with on-the-fly EUP on a more general level, while Themes 4--5 pertain to \textit{usability} aspects of Tabula's specific implementation. For each theme, we present its definition and use participant quotes to provide support. Theme~4 is further organized into subthemes to more explicitly illustrate its different facets. Participant quotes are attributed by participant ID, with minimal edits made to ensure clarity while retaining meaning.

\subsection{\textbf{Theme 1: End users viewed program steps to better understand the system}}
%Our analysis results reveal that the user's behavior of checking the review mode is followed by the user wanting to confirm their expectations about the program or is used to provide them with insights on how the program actually "understood" the user's intent. 
The first theme captures how users reflect on system rules when viewing program steps (see Figure \ref{fig:system}, Right), initiate revisions to the program after discovering an error, and proactively utilize the program steps to make programs incrementally. Reflections shared by five participants highlight the end users' heavy reliance on step visualization when understanding system operation.

\paragraph{Helping in recalling system rules} Five participants recalled key system rules as they connected those rules to shaping expectations and interpreting the final output of the program steps. Participants explicitly stated concepts such as \inlinequote{recordings} (P6), \inlinequote{loops} (P6), and triggers, \eg{} \inlinequote{adding [my] stops} (P15). Similarly to P6's comment on recordings, P10's comment on the number of recordings provides insight on how users were able to evaluate system output as they viewed program steps and remembered key concepts. P10 mentions, \inlinequote{I think that was... yeah I don't know what the third [recording]'s supposed to be} as they viewed the program steps. For the case of recalling how to specify program logic, P6 asked a critical question \inlinequote{does it loop?} as they meticulously viewed each program step.

\paragraph{Initiating revisions after discovering errors} Besides the phenomenon \finalfix{described in the previous paragraph}, there were instances where users motivated themselves to match the program steps provided by the system to the steps they imagined and desired. When misaligned, end users took the initiative to redo the entire program or wanted to make a revision after they viewed the steps. Six participants felt they wanted to \inlinequote{do it again} (P12) as they examined the final steps. P9 displayed confidence as they noticed what output of the program steps were \inlinequote{obviously [...] wrong} and expressed the urge to redo it by saying \inlinequote{because I know exactly why.} P21 expressed a related sentiment and wanted to make partial revisions, as they cited \inlinequote{so I go to just delete this recording.} Participants were able to make these reflections and express their urge to revise the program because they viewed the detailed program output.

\paragraph{Incremental revisions} In addition to the more common ways participants interacted with the visualized program steps, P15 also used the program steps to create programs incrementally. Deliberately checking the review panel and the detailed output of the program steps, P15 planned for the next recording after citing \inlinequote{okay let me see what it did here this is only one of my things.} Additionally, revision of the \inlinequote{next instruction} was made as they have continued citing \inlinequote{is not when someone says stop but when someone says go to when someone says where is [the] kitchen.} This observation brings us insight into the importance of including detailed step visualizations within the system, rather than solely focusing on how to capture user intent with regards to system rules.

\subsection{\textbf{Theme 2: End users have poor mental model of the input paradigm}}
This theme reflects end users' experiences with the non-sequential, rapid specification of the robot programs. Instead of \finalfix{requiring} \textit{step-by-step} instructions, Tabula accepts \textit{non-sequential} input\finalfix{, }\ie{} end users need not instantiate commands and locations in the exact order to be performed. With Tabula's non-sequential input, users provide a verbal task hint and the sketch on the tablet interface, and then these inputs are synthesized into a list of program steps by the system. Twelve participants indicated that the way that the interface required them to provide input was unintuitive. %This misalignment affected their perception of the programming tool as well as their comfort and ease of using it.

%\paragraph{Difficulty with system-specific rules}
%Six participants specifically felt that the usage was \inlinequote{rigid} (P2, P5) because \inlinequote{the rules are kind of strict} (P10). Some of this rigidity stems from the specific implementation of the Tabula system, such as not being able to lift the finger from the tablet while drawing a sketch. P11 shares their experience with this restriction, saying that when they felt they had made a mistake, they immediately \inlinequote{threw their finger up because it felt natural.}

\paragraph{Preference for step-by-step inputs}
Eleven participants articulated that they would have preferred the flexibility to interchange the speech and sketching inputs, especially to support specifying programs step by step rather than non-sequentially. For example, P16 felt that first giving the speech input was \inlinequote{backwards} when specifying a task for the robot to greet patrons at the front of the store because they want to \inlinequote{first get [the robot] to the entrance and then give the command.} P14 expressed similarly that for them, it was easier to draw out the path and then think of the speech because they first need to \inlinequote{invite [their] mind to imagine that place} then provide the speech command. 
The comments from these participants evoke the sense that they are thinking of the robot program in a step-by-step, linear sequence of actions, which contrasts the non-sequential input paradigm used by Tabula. Other participants were more explicit about their preference for \inlinequote{step wise} (P3) inputs and found it \inlinequote{difficult} (P19) to adapt to the non-sequential pattern. P17 explains how they would have preferred to create the robot program that brings the groceries to the kitchen, saying:

\begin{quote}
    \inlinequote{The robot goes to the garage, and then [I'll] tell them `Take the groceries.' I'll put the groceries to the kitchen, and [I] draw [a path] to the kitchen.}
\end{quote}

Overall, these participants seemed to struggle with the misalignment between their step-by-step mental model of the robot program and the non-sequential inputs they were asked to provide.

% possibility for contradicting -- P18 --> didn't really fit after all
% \begin{quote}
%     \inlinequote{yeah so if I say something wrong like people always make mistake when they saying something like I want to move to the I say I want to move to the bedroom I make a mistake because I really want to move the bathroom that's why I draw to the bathroom so there's a conflict with my  saying because I will change my mind it's really quick thing} (P18)
%     %yeah so if I say something wrong like people always make mistake when they saying something like I want to move to the I say I want to move to the bedroom I make a mistake because I really want to move the bathroom that's why I draw to the bathroom so there's a conflict with my  saying because I will change my mind it's really quick thing (P18)
% \end{quote}

\paragraph{Unintuitive to program robot remotely}
One remaining participant articulated that they did not like creating a robot program when the robot was not in the same location. They expected to \inlinequote{let the robot come to [them] first and then give [the robot] a task} (P17) instead of using the tablet interface to do so remotely. While only one participant expressed such a differing model of creating programs for the robot, it highlights a different aspect to the input paradigm that was not widely explored in this study.

% supporting work of input mental model alignment being important:
% JESSIE (through its decision of what to expose to users)
% Polaris (through the deliberation between actions and goals)
% Rao Kambhampati's work on mental model alignment

% Implication -- users prefer step-by-step instructions.
% Recommendation -- Feedback should be geared towards mental model alignment, as in Rao's work. Alternatively, on-the-fly interfaces can be designed with users' mental models in mind.  

\subsection{\textbf{Theme 3: End users felt that the robot was a tool to use}}
End users viewed creating the robot programs as a way to utilize a robot like a tool rather than treating the robot as an independent, autonomous agent capable of reasoning about its environment. This theme is formed from seven participants' remarks, and it encapsulates a unique way in which they viewed the robot. Based on the demographic data, we see a potential relationship between experience with programming
languages and how people perceived the robot---of the seven participants who 
reported no experience with programming languages, only one of these participants articulated the robot was a tool rather than an autonomous agent.

\paragraph{Learning to use the tool}
Participants felt it was necessary to learn the specific rules to use Tabula because \inlinequote{if you buy anything you want to use you have to read and use the manufacturer's manual to be able to understand how to use it} (P20). This viewpoint emphasizes the robot's role as a product to purchase and use as a tool.
%Although the robot programming process required learning specific rules, participants felt that it was a \inlinequote{great investment} (P2) because \inlinequote{if you buy anything you want to use you have to read and use the manufacturer's manual to be able to understand how to use it} (P20). This viewpoint emphasizes the robot's role as a product to purchase and use as a tool.

\paragraph{Prioritized the robot's capabilities over their own preference}
Two participants built on the notion of learning the robot's specific rules by indicating that as they created their robot programs, they prioritized adapting their inputs based on their perception of the robot's abilities. P3 described that while \inlinequote{it was easy enough for [them] to do one thing or the other,} they opted to create robot programs based on \inlinequote{whatever [they] thought it was easier to implement for the robot.}

\paragraph{Need to ensure real world matches robot's world model}
In a more extreme view, four participants felt they had a direct responsibility to ensure that the reality reflected the assumptions that the system made because the robot would not have the reasoning capabilities to troubleshoot deviation. P8 articulates this point clearly, saying:

\begin{quote}
    \inlinequote{If [the robot] assumes that [a container is] going to be there, then it's your responsibility to make sure that [...] the containers [are] there to for the robot to put [the object] in.}
\end{quote}

This perspective shifts the responsibility onto the end user to ensure that the robot is able to succeed at its program, rather than expecting the robot to reason about the world autonomously. 

% \paragraph{Takeaway?}
% It also requires the end user to develop a deeper understanding of the robot's capabilities and model of the world so that they can have the requisite knowledge to ensure the robot's success. 

% \paragraph{Relation to demographics??}
% Based on the demographic data, we see the potential for experience with programming languages to impact how people perceived the robot's role as an extension of themselves---in total, seven participants reported no experience with programming languages, but only one of these participants expressed the sentiment that the robot was an extension of themself. Future research is needed to correlate demographic background with perceived/desired role of a robot to inform future research.

\subsection{\textbf{Theme 4: End users had mixed experiences on interaction with the program synthesizer}}
End users either appreciated that the program synthesizer provided human-like common sense support, or they disliked the assistance because they perceived it as a loss of control over the robot program. Participants interacted with the assumptions made by the program synthesizer when it automatically inserted missing actions and objects. Twenty participants specifically commented on this aspect of the system, revealing a dichotomy of end users who either appreciate or reject the notion of the program synthesizer making automated assumptions and a small subset who had mixed perspectives. The two subthemes presented below illustrate the two prevalent, opposing viewpoints of the system. 
Based on the demographic data, we see the potential for experience with programming languages to impact how people perceived the interaction with the program synthesizer---of the seven participants who reported no familiarity with any programming languages, only one participant appreciated the automated assumptions. %Participants who were familiar with programming languages were split between the two groups, but without further information about their level of expertise, we are not able to infer further details about the relationship between experience with programming language and experiences on interacting with the program syntehsizer.

\subsubsection{Subtheme 4a: Automated assumptions can offer support to end users}
The 11 participants who spoke positively of the automated assumptions made by the program synthesizer expressed that it was \inlinequote{natural} (P12) and that it provided support during their programming experience. 

\paragraph{Interactions were more natural/human-like}
Participants specifically commented on the assumptions made about when to insert an object and when to insert actions, indicating that these assumptions offered a desirable level of human-like common sense from the robot that is \inlinequote{helpful} (P5). For example, P18 expressed that the actions automatically inserted by the program synthesizer simplified the process for them because \inlinequote{the put action combined the grab and the move and stuff} which meant that they did not need to take time to think through or add those actions--that burden was offloaded to the program synthesizer. 

\paragraph{Desire for additional automated support}
Four participants further indicated that the system could be more helpful by making additional assumptions based on user input. For example, P4 wished that the system would automatically generate a condition for \inlinequote{exiting the loop}, while P13 wanted the system to \inlinequote{provide suggestions} if the user made a mistake. P11 further envisioned that the system could make assumptions based on the robot's ability to interact with objects that it is close to in its environment, such as \inlinequote{if you move [the robot] to the item, [the system] just infers that [the robot]'s supposed to pick it up.} 
The automatic assumptions provided convenience to some end users, who felt support from the system for easing into the robot programming process.

\subsubsection{Subtheme 4b: Automated assumptions can lead to loss of control}
The 13 participants who commented negatively about the use of automated assumptions felt that these assumptions led to a loss of their ability to control how the robot would act. Eight of these participants focused comments on the automatic insertion of objects and items, while the remaining 5 participants expressed the desire for more control over the robot's precise movements within its environment, such as indicating specific regions to avoid. 

\paragraph{Doubting robot's knowledge to automatically insert objects/actions}
Eight participants focusing on the automatic insertion of objects and actions felt that it was \inlinequote{unnatural} (P2) and questioned whether the robot could or should have enough knowledge of the environment to make such assertions. For example, P2 felt that depending on the scenario, the user may or may not intend for an object to be placed inside of a container. P2 explains:

\begin{quote}
    \inlinequote{Given certain use cases, I could imagine like if you have one of these robots moving gravel around a yard, you probably wouldn't have a container there, but uh in [the grocery delivery] scenario it felt right to assume that there would be a cabinet.}
    %I guess there doesn't necessarily need to be a cabinet there, and get given certain use cases I could imagine like if you have one of these robots moving gravel around a yard, like you probably wouldn't have a container there, but uh in this scenario it felt right to assume that there would be a cabinet
\end{quote}

From P2's example, it may be difficult to infer when an object should be placed in a container or not. P6 similarly felt that it was reasonable for the end user to have to explicitly specify whether there is a container, saying \inlinequote{It just makes sense if I have to tell it that there's a teddy bear in the middle of the floor that I should also have to tell it that there's a cabinet on the wall.} P19 echoes the sentiment that they \inlinequote{don't know exactly what [the robot]'s going to assume to do and especially with the assuming where it's going to put.} Overall, participants who did not like the automatic insertion of objects and actions felt that they did not have as much control or understanding over how the system would behave.

\paragraph{Desire to control robot's location}
Five participants viewed the abstraction of the environment into regions as opposed to exact coordinates as a negative assumption of the system---they wanted more control over the precise location or path the robot would travel within the space. The current system abstracted away precise coordinates in favor of general semantic regions such as \inlinequote{kitchen} or \inlinequote{living room.} P20 explains their desire, using the example that the kitchen is a \inlinequote{big place [...] so maybe [in] the command there should be a way to specify where exactly in the kitchen you want the groceries to be placed.} Building off of this sentiment, P5 and P13 both expressed that they may want the robot to \inlinequote{avoid a certain area} (P5), so the path that they draw for the robot is the precise one that it should follow. This group of participants includes the four participants who also spoke favorably about the automatic assumptions regarding actions and objects in Subtheme 4a, which indicates that there is a need to create a balance between easing the programming process and giving the users the desired level of precision over the robot's behaviors.

\subsection{\textbf{Theme 5: There is more to using the system than understanding its basic functionality}}
Even after understanding Tabula's basic functionality, end users still faced a learning curve to become proficient. 

\paragraph{Translating rules into use}
Eleven participants expressed that there were \inlinequote{differences between understanding and doing} (P12). P10 articulated that rules for creating robot programs led to instances where \inlinequote{you have something in your mind but you don't know how to immediately put it in the system.} This \inlinequote{gap} (P12) forced P15 to resort to \inlinequote{taking different parts of the training and kind of consolidating it into doing a scenario.}

\paragraph{Performance aspect to making recordings}
In addition to conceptual difficulties with \inlinequote{connecting the dots} (P15) between various concepts, participants also noted that making the recordings created \inlinequote{a performance aspect [...] to get it all in one go} (P2). Once participants began a recording, they had to \inlinequote{remember the vocabulary that the robot would understand} (P10). If they made a mistake or if the system \inlinequote{had a hard time} (P4) discerning what participants said, then they had to delete the recording and start again.

\paragraph{Desire for editing support}
While some participants seemed comfortable with the iterative process of creating, reviewing, deleting, and re-doing recordings, others wanted a different way to correct mistakes. Four participants wanted the ability to \inlinequote{edit a recording afterwards} (P7), which would ease the pressure of providing precisely correct speech and touch on the first attempt. Two participants wanted a quick way to \inlinequote{erase if you messed up} (P11) during a recording without having to \inlinequote{restart} (P8) the whole program.

% \paragraph{Takeaway?}
% Overall, we had anticipated that learning the rules of the interface would be the biggest challenge facing the participants. However, instead, it seemed that for some participants, the bigger challenge came from applying the rules or successfully providing the intended inputs.

% Supporting work
% 

\section{Discussion}
We sought to better understand end user experience with on-the-fly robot programming through an in-depth assessment of the open-source EUP tool \textit{Tabula}. Through our investigation, we uncovered themes that provide insight into various aspects of on-the-fly robot EUP. Some themes relate specifically to the implementation of Tabula, such as its use of multimodal inputs and the use of automated assistance in the form of a program synthesizer. However, combined with prior work, other themes point to broader implications regarding the concepts realized through Tabula, such as the reliance on the visualized program steps. Overall, we see the promise of on-the-fly EUP tools as a way to facilitate the use of robots to aid with day-to-day tasks, but these tools require further research and refinement before they will be sufficient. We encourage future researchers to conduct more in-depth user studies with existing or novel EUP tools so that we can build a better understanding of end user needs based on a variety of on-the-fly EUP tools.

\modified{In the paragraphs below, we provide general points of discussion of our findings, such as how participants perceived the role of the interface and how our findings relate to the Cognitive Dimensions of Notation (CDN) \cite{green1996usability}. We reserve detailed discussion of implications for future design of EUP systems for \S\ref{sec:implications}, highlighting four key design recommendations.}

\modified{\paragraph{Role of the interface} 
%We note that some participants equated the interface's automated assumptions to the robot's level of autonomy. 
We note that participants had different assumptions about the role of the interface, with some thinking that its capabilities were limited to capturing user input on behalf of the robot and others attributing planning and reasoning capabilities to the interface itself. 
Specifically, some participants felt that the robot was directly generating the sequence of steps. For example, in Theme 3, P3 discussed \inlinequote{If [the robot] assumes that [a container is] going to be there...}), which attributes the automatic assumption made to the robot rather than to the interface's underlying synthesizer. Also in Theme 4-Subtheme 4b, P2 spoke as though the \emph{robot} was assuming that there was a cabinet to place the groceries within the environment. Therefore, we believe that some participants attributed the automated decision making to the robot's autonomy rather than the features of the interface.} 

\modified{We find this connection particularly interesting, considering that there was no physical robot present during the study. The distinction between the EUP tool and the robot's autonomy is blurry, especially because the EUP tool may depend on specific robot capabilities. Given that participants did not necessarily separate the EUP tool from the robot's capabilities, we can consider the automated decision making of Tabula in close alignment with the autonomy of the robot for which it was being used to generate programs.}

\modified{
\paragraph{Relation to Cognitive Dimensions of Notation (CDN)}
CDN is a set of 14 design principles intended for evaluating programming languages, notations, and user interfaces \cite{green1996usability}. Each principle illustrates one aspect of usability, intending to serve as a guide toward improving usability along specific dimensions. We found that two dimensions align particularly well with certain aspects of our themes, indicating that these dimensions are key to future robot EUP tools. While CDN is helpful in contextualizing usability aspects of Tabula, CDN does not explicitly discuss autonomy or perceptions surrounding interaction with automated decision making.
}

\modified{
The first dimension, \textit{progressive evaluation}, considers how easily users can evaluate and obtain feedback. This dimension connects well to Theme 1. Participants largely relied on the generated list of program steps as the mechanism for receiving feedback and updating their solutions accordingly, indicating that supporting progressive evaluation is critical.
}

\modified{
The second dimension, \textit{premature commitment}, considers both how strong the constraints are on using the system and also how users can easily change or correct decisions later on. With Theme 2, participants felt constrained by the strict order of speech and sketching inputs. Given that participants had varying notions on the best input order, avoiding premature commitment by providing more flexibility in speech and sketch inputs is crucial. Theme 5 also supports the need for premature commitment as participants wanted a way to edit the recordings after the fact instead of having to delete and re-do them.
}

%\modified{
%Other dimensions relate less strongly to specific findings within themes. For example, \textit{error-proneness} considers the extent to which the system's notation can influence the likelihood of mistakes---participants in Theme 5 expressed that the interface had a \inlinequote{performance} aspect which caused them to make mistakes. Participants in Theme 5 also expressed the difference between understanding the rules and being able to use the system effectively. While this experience does not perfectly align with the concept of \textit{hard mental operations} (\ie{} how easily participants are able to keep track of what is happening), we can interpret their difficulty as a symptom that Tabula's current implementation relies on hard mental operations.
%}

%Theme 1 - progressive evaluation, users were achieving progressive evaluation through checking the generated steps
%Theme 2 - premature commitment, users had to provide all of hte recording input before they could receive any feedback, and the strict order of inputs posted problems
%Theme 3 - not really covered
%Theme 4 - subthemeA, subthemeB not super supported, although perhaps there's a relation to role-expressiveness
%Theme 5 - not entirely, although translating rules to use hints at error-proneness and desire for editing support points at prematuer commitment

\begin{figure*}[!bt]
    \centering
    \includegraphics[width=\textwidth]{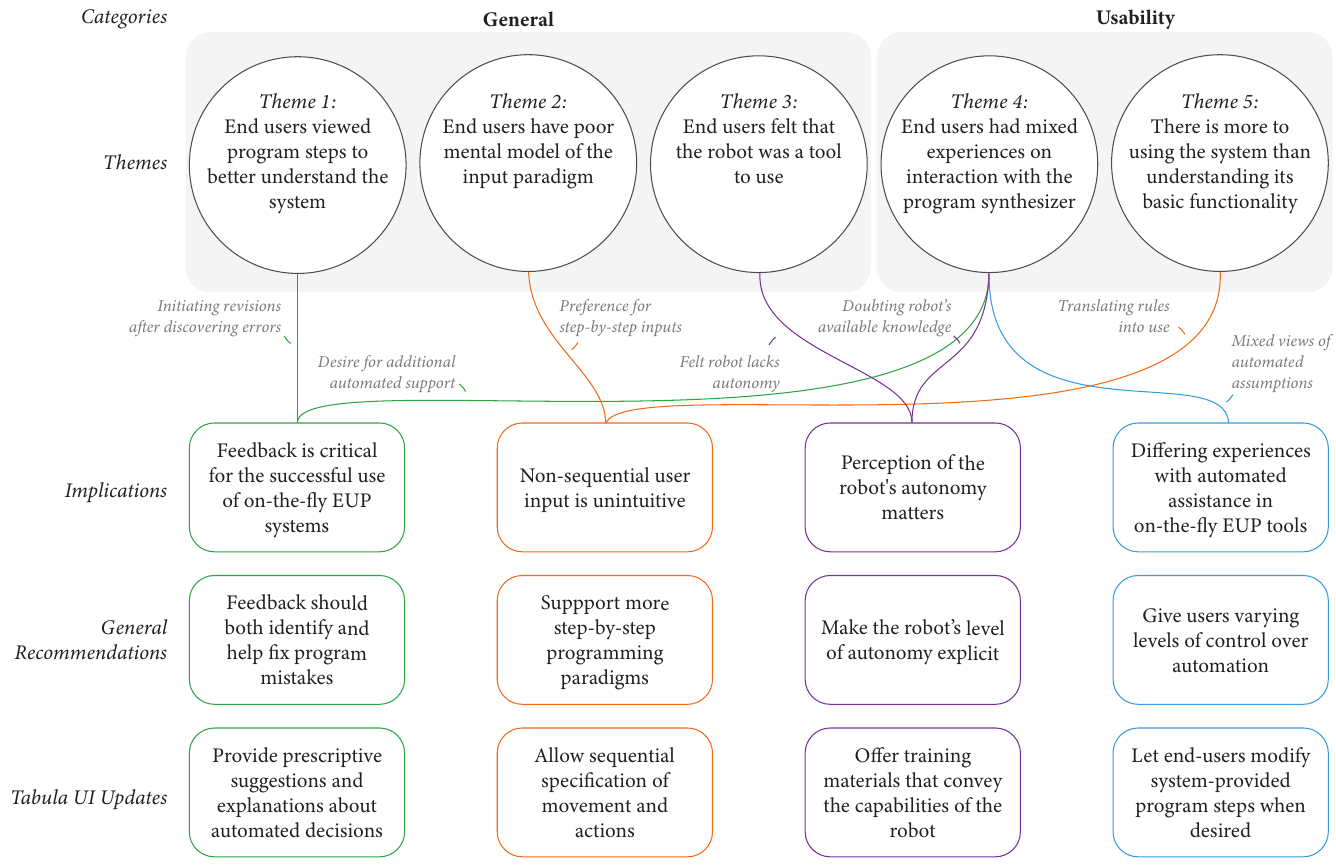}
    \caption{An overview of the connection between the findings and resulting design implications and recommendations.}
    \Description{Graph with three rows:

The first row has 5 circles, each with the texts:
1. Theme 1: end users viewed program steps to better understand the system
2. Theme 2: End users have poor mental model of the input paradigm
3. Theme 3: End users felt that the robot was a tool to use
4. Theme 4: End users had mixed experiences on interaction with the program synthesizer
5. Theme 5: There is more to using the system than understanding its basic functionality
Circles 1, 2, and 3 are labeled with the category General, and Circles 4 and 4 are labeled with the category usability.

The second row has 4 boxes labeled as Implications, each with the texts:
1. Feedback is critical for the successful use of on-the-fly EUP systems
2. Non-sequential user input is unintuitive
3. Perception of the robot's autonomy matters
4. Differing experiences with automated assistance in on-the-fly EUP tools

With connections from circles in row 1 to boxes in row 2:
1 to 1: Initiating revisions after discovering errors
2 to 2: Preference for step-by-step inputs
3 to 3: Felt robot lacks autonomy
4 to 1: Desire for additional automated support
4 to 3: Doubting robot's available knowledge
4 to 4: Mixed views of automated assumptions
5 to 2: Translating rules into use

The third row has 4 boxes labeled as Recommendations, each with the texts:
1. Feedback should both identify and help fix program mistakes
2. Support more step-by-step programming paradigms
3. Make the robot's level of autonomy explicit
4. Give users varying levels of control over automation
Each box in row 3 is directly under the corresponding numbered box in row 2.

The fourth row has 4 boxes labeled as Tabula UI Updates, each with the texts:
1. Provide prescriptive suggestions and explanations about automated decisions
2. Allow sequential specification of movement and actions
3. Offer training materials that convey the capabilities of the robot
4. Let end-users modify system-provided program steps when desired
Each box in row 4 is directly under the corresponding numbered box in row 3.}
    \label{fig:discussion}
    %\vspace{-9pt}
\end{figure*}

\subsection{Design Implications}\label{sec:implications}
Based on the themes discussed in \S\ref{sec:findings}, we present design implications and recommendations to inform future design of on-the-fly EUP tools. \modified{Each recommendation includes a general recommendation of how the implication could be applied generally to EUP tools, as well as a specific suggestion for modifications which would lead to ``Tabula 2.0.''} The link between the findings, implications, and recommendations is visualized in Figure~\ref{fig:discussion}.

%The previous, long version: Visual feedback of the program steps should be closely coupled to user inputs. Visualizing complex programs to novice end users should be more closely explored in future research, particularly when considering how to display more complex logic.
\subsubsection{Design Implication 1: Feedback is critical for the successful use of on-the-fly EUP systems.}
Our findings highlight the importance of integrating feedback mechanisms within on-the-fly \modified{EUP} systems like Tabula---in contrast to tools in which program flow is explicitly embedded within user input (\eg{} block-based programming tools \cite{chung2016iterative, coronado2021towards, beschi2019capirci}), Tabula users rely heavily on feedback \modified{(\ie{} program step visualizations)} to understand system behavior and make program changes (Theme 1). 
\modified{Even without access to a way to deploy their programs to a simulated or physical robot, participants were still able to use the step visualizations as a pre-deployment check as a way to understand where an error occurred and how they could adjust the program flow to correct it.}
% \modified{These results also indicate that visual feedback can still be beneficial without having a physical robot performing a task and providing real-time feedback. Visualizations of program steps can act as simulations pre-deployment and provide end users on where and why an error has occurred and how that could modify the program flow.}
%add other examples of existing EUP tools?

The reliance on feedback echoes prior investigations of end-user developers interacting with a program synthesizer \cite{porfirio2019bodystorming}. \textit{How} feedback is applied is additionally crucial to human-AI systems in general \cite{amershi2019guidelines}, and our results suggest that purely \textit{descriptive} (as opposed to \textit{prescriptive} or \textit{explanatory}) feedback can lead to a lengthy process of discovering system behavior. \modified{Specifically, because participants were only provided with the resulting program steps (descriptive feedback), they had to use their own judgment to discern if their program was correct and guess how to modify their inputs to Tabula in order to achieve the desired program output (Theme 1). }In Theme 4-Subtheme 4a, participants specifically expressed that the system could provide additional automated support to further ease their programming efforts. \modified{This additional support included prescriptive measures such as the system detecting mistakes and offering corrective suggestions and preemptive measures such as the system generating conditions on their behalf (P4 and P13).}

\paragraph{Recommendation 1: 
%\modified{On-the-fly EUD tools should provide visual feedback and also offer} users not only information on what is wrong with a program but also information on how to fix it or explanations for why the system behaved in a certain way.} 
For on-the-fly EUD tools, \modified{visual} feedback should provide users not only with information on what is wrong with a program, but also with information on how to fix it or explanations for why the system behaved in a certain way.}
On-the-fly EUD should therefore draw from prescriptive approaches in formal methods, such as proposing repairs \cite{chung2020iterative}, and strengthen its descriptive approach through explainable AI techniques such as model reconciliation \cite{chakraborti2019plan}. \modified{Approaches such as these could offer end users the ability to assess incomplete solutions, obtain feedback, and build programs based on the interface's suggestions.}

\modified{Specifically, in realizing a ``Tabula 2.0,'' we can clearly label steps provided directly by the end user and steps generated by the system. Then, the end user could select steps and ask a question such as ``Why is this step before that step?'' Using methods such as iterative planning as outlined by \citet{smith2012planning} or \citet{wang2024describe}, the system can interactively offer a rationale behind the decision and suggest new constraints to add or modifications to existing recordings which would alter the resulting steps.}

\subsubsection{Design Implication 2: Non-sequential user input is unintuitive.}
We found across Themes 2 and 5 that participants struggled with the rules and input paradigms that Tabula enforces. Participants had to provide input following a strict pattern and adhere to usage rules, which they expressed resulted in feelings of frustration because they could not easily use the interface to express their intention. \modified{For instance, participants commented on the \inlinequote{backwards} input paradigm that Tabula enforced and conveyed their preference for \inlinequote{step-wise} inputs (P3, P14, and P16).} This study included a fairly extensive tutorial which included many interactive examples, yet it is evident that additional training would be required for participants to achieve proficiency. The underlying representations of user input and resulting program steps appear to be critical to Tabula users' experience, a finding that aligns with prior work of user program comprehension---certain program representations may align better (\ie{} representations that facilitate forward-reasoning \cite{trafton1991providing}) or worse (\ie{} the imperative programming paradigm \cite{kambhamettu2021collecting}) with user intuition. Other representations are prone to misalignment between user mental models of program behavior (\ie{} trigger-action programming \cite{huang2016trigger}) or may result in reduced user performance (\ie{} visualizations of data flow rather than control flow \cite{good1999vpls}). Fortunately, motivated by prior work that uses formative evaluation to inform product design \citep[\eg{}][]{li2019pumice}, we believe that changes to the interface can improve user experience with non-sequential input, such as through the inclusion of the ability to edit recordings after they are created. Therefore, it will be important to balance efforts to create intuitive tools for end users with developing effective training protocols for introducing new paradigms and systems.

\paragraph{Recommendation 2: Find a way to design on-the-fly EUP tools that supports more step-by-step programming paradigms.} Training to use on-the-fly EUP tools to create robot programs should remain essential, even if the training eventually becomes teaching end users about a robot's capabilities and limitations. However, interfaces can always be designed to be more intuitive, \eg{} by supporting more step-by-step paradigms where users can specify movement and actions sequentially, through methods such as participatory design and research through design. 

\modified{In realizing ``Tabula 2.0,'' we would remove the restriction imposed during our study of speech needing to occur before the sketch, update the synthesizer to allow end users to link utterances to specific waypoints, and add further support to accommodate multiple, separate speech utterances per recording. In making the above modifications, end users will have more flexibility and control to be able to specify step by step what the robot should do at which location. The result will be a system which would allow participants to interleave sketching and speech, similarly to how tools like \textit{Figaro} \cite{porfirio2021figaro} allow more sequential specification of movement and actions. Unlike Figaro, however, the system would still insert or complete missing or incomplete specifications.}

% related work to be integrated later:
% - Collecting Insights about How Novice Programmers Naturally Express Programs for Robots
% - PUMICE is an example of a system designed from collected insights
% - Providing Natural Representations to Facilitate Novices' Understanding in a New Domain: Forward and Backward Reasoning in Programming ("forward reasoning" is more intuitive for people)
% - VPLs and Novice Program Comprehension: How do Different Languages Compare? ("visual representation matters")
% - The right feedback can also be critical here (cite Rao's work)

\subsubsection{Design Implication 3: Perception of the robot's autonomy can either limit or enhance the role of the robot as a collaborative entity.}
Theme 3 and Theme 4-Subtheme 4b together illustrated that a subset of participants perceived that the robot was not necessarily able to reason about its world. Participants from Theme 4-Subtheme 4b expressed this view through distrust of the automated assumptions of the program synthesizer \modified{(\eg{} P2 considered the automated assumptions to be \inlinequote{unnatural})}, whereas participants in Theme 3 felt that the robot was merely a tool to use \modified{(\eg{} P20 felt there would have to be a \inlinequote{manufacturer's manual} such as the instruction booklets that come with other household tools)}. Misperceptions of robot capability \cite{cha2015perceived} and the potential to view the robot as a \inlinequote{tool} (rather than having agency) \cite{takayama2012perspectives} are known phenomena in human-robot interaction. Our interviews not only suggest that these phenomena translate to EUP, but also that, in participants' words, user perception of the robot's autonomy changes their behaviors and experiences. 
We further saw that in both of these themes, the participant's prior familiarity with programming languages may have been a factor impacting their current perception of the robot's autonomy. Given that a robot's level of autonomy may be set, it is important to think about how to communicate the robot's level of autonomy and precise role to the end user.

%in LoRA (Table 4)- https://www.ncbi.nlm.nih.gov/pmc/articles/PMC5656240/, authors specify either the human or the robot doing the sensing, planning, and acting
%could provide examples on 'shared control with (xx) initiative (where either the human or the robot take initiative on the planning)' with supervisory control (where robots do the planning and the human monitors with an overriding capability for plan revision)

\paragraph{Recommendation 3: When designing an EUP tool, the robot's level of autonomy should be made explicit.} Tools designed for autonomous robots who can reason about their world may differ from tools designed for using robots to extend human abilities. \modified{Tabula was designed with the intention that the system/robot could reason about the world, such as understanding when it may need to automatically insert steps or assume that certain objects would be present (\eg{} assuming the kitchen cabinets are there to put the groceries in). However, as some participants did not appreciate this level of autonomy of the system, they desired more low-level control over specifying exactly what to do at which locations. Future research should explore these differences, such as by investigating ways in which different levels of autonomy and agency can be communicated to end users. }%, \eg{} by providing end users with exemplary interactions that they can program with the robot while leveraging different levels of robot autonomy. }
% \modified{Adjusting to different levels of robot autonomy and user control can be instantiated by both entities sharing control (\eg{} the synthesizer can automatically complete the user's command while the user has control on which options to implement) or the end user taking \textit{supervisory control} over the robot-provided plans (\eg{} the user may redo the operations that the synthesizer implemented and treats it as a decision support system)\cite{beer2014toward}.} %Future research should explore these differences, such as by providing additional instruction and experience with the target robot platform prior to evaluating the specific EUP tool. 
While it is not necessarily the case that each specific robot with varying autonomy levels would require a different EUP tool, \modified{different robot characteristics will likely indicate the need for more specific EUP tool features. Incorporating a conceptual framework such as the robot autonomy scale (see \citet{beer2014toward}) could create more transparency with regard to how much automated support is provided to the end user, and such integration will therefore be a necessary step for future EUP tool design.}

\modified{For ``Tabula 2.0'', we can clearly situate Tabula's level of autonomy within the scale of robot autonomy \cite{beer2014toward} as \textit{sharing control} (\eg{} the synthesizer can automatically complete the user's commands while the user has control on which commands to instantiate). We can specifically communicate the sensing, planning, and acting capabilities of the robot that is connected with Tabula through training materials that exhibit specific use cases of the synthesizer. 
Within the training materials, we will emphasize the exact capabilities the robot has, along with the extent to which the synthesizer makes assumptions about the user's intent. This training will be particularly critical for those who are not familiar with robots, although as familiarity with robots increases, the need for in-depth training will likely taper.
}

%is implemented through a high system autonomy for all function allocations (\ie {} sensing, acting, and planning) yet not fully instantiated within many interactions that the user could have with the system, }

% Implication -- view of robot as "tool" limits the robot as a collaborative entity.
% supporting work
% Social robots - beyond tools to partners (viewing the robot as not just a tool but a collaborator)
% The "artificial" colleague -- we view AI collaborations as less meaningful than human-human collaborations, and view AI as a subordinate
% Theory of social agency for HRI -- Black Jackson and Tom Williams
% Recommendation -- incorporate HRI theory of robots as partners into the design of EUD systems

\subsubsection{Design Implication 4: As demonstrated with Tabula, user experience with perceptions of automated assistance varies with on-the-fly EUP tools.}
%observation
Particularly with Theme 4, we observed that end users were split between appreciating the support of the automated assistance provided by the program synthesizer and wishing that they had more control over the programs generated. A handful of users expressed mixed opinions. These varying preferences may be due to our participant pool including a diverse background of programming, video game, and engineering experience. We note that robots in the home will similarly be used by individuals with varying backgrounds. Prior work in robot EUP acknowledges the need to cater to varying backgrounds by providing entry points for different types of developers \cite{huang2017code3, glas2012interaction, pot2009choregraphe}; our work suggests that in addition to providing multiple entry points,
%. Tools that are either designed to support novice or more experienced users, but not both, may not be generally accepted and marketable. Instead,
on-the-fly EUP tools will need to cater to a sliding scale of preferences regarding the level of user control versus automated assistance.

\paragraph{Recommendation 4: When incorporating automation, also give users varying levels of control.} 
Ideally, anything that is handled by some form of automated assistance should also be directly controllable and/or modifiable for the end user. However, \finalfix{in reality,} due to \finalfix{Tabula's} non-sequential multimodal inputs, all aspects of \finalfix{its} automated assistance may not be fully customizable. Therefore it is critical that EUP tools clearly communicate explanations behind their decisions and offer guidance to support end users in creating the programs that they desire.

%I think we can be more specific. For human-in-the-loop, synthesizer will allow people to prioritize or re-arrange program steps? For low autonomy/high control, people can manually add steps to the generated program?
\modified{In continuation of the discussion on robot autonomy and end-user control, ``Tabula 2.0'' can further be designed to allow end-users to specify their preferences with regard to the program synthesizer. Users should be able to adjust the level of control wielded by the synthesizer, in particular the degree to which the synthesizer involves the human in the loop. For instance, users who opt for high level of control and low level of robot autonomy should be able to prioritize or re-arrange the program steps suggested by the synthesizer.} 

% see guidelines from Amershi et al.

\subsection{Limitations and Future Work}
Our work has a number of limitations that point to future work. \modified{We separate our limitations and future work into three categories.}

\paragraph{\modified{Generalization to EUP}}First, our investigation of end-user experience with on-the-fly robot EUP used one existing multimodal EUP tool with specific capabilities and constraints. \modified{Combined with the fact the on-the-fly development paradigm remains novel for EUP tools within HRI,} the extent to which the behaviors we have observed and the perceptions we have documented will generalize to existing EUP tools is unclear. \modified{With that being said, the purpose of our investigation is less to understand existing EUP tools, but moreso to uncover guidelines for designing EUP tools within this novel paradigm. Future work must therefore apply our recommendations to the design and evaluation of a ``Tabula 2.0.''} 
\modified{Future work must also extend our investigation to additional tools that represent different EUP approaches, including input methods, intent inference, and program generation methods so that we can understand what is unique about the multi-modal on-the-fly paradigm and what generalizes to robot EUP as a whole.}

\paragraph{\modified{Learning Tabula}}Second, although the EUP paradigm is intended to be accessible to non-experts in programming and robotics, effectively using any complex end-user tool requires learning and gaining comfort with its use, which put limits on how long our participants could explore the tool as well as the time needed to generate programs. \modified{Future work should include multi-day field studies with Tabula to investigate how real users learn to use, gain familiarity with, and generate or refine programs using on-the-fly EUP tools, similarly to the approach taken by \citet{ranganeni2024robots}. Additionally, recent work in tablet-based EUP suggests that end-user perceptions, experience, and success using EUP tools is tied to individual background \cite{porfirio2024polaris}.
%Our investigation finds evidence for a potentially relevant phenomenon---appreciation for Tabula's automated assumptions was lacking amongst participants who shared a common background characteristic, namely lack of programming experience (Theme 4), but it remains unclear whether a link between individual technical expertise and Tabula usage exists.
Future work with Tabula should therefore investigate possible links between relevant user background characteristics and tool usage in order to better inform strategies for training Tabula end users.}

\paragraph{\modified{Involvement of a Robot}}Third, participants used the EUP tool in a sandbox and therefore did not see their programs being executed on an actual robot. Their programs were only represented on the EUP tool. The ability to see the robot behaviors that result from their programs might provide participants with stronger mental models of the capabilities and limitations of the robot platform they are working with, which might inform their programming choices.  

\modified{
\paragraph{Exploration of Specific Application Domains}
Finally, our evaluation engaged the general population of our campus community. While this population could represent home robot users, it does not represent specific domain experts such as shop keepers or healthcare workers. Each domain may have its own needs which dictate how end users would perceive Tabula. We are currently exploring applications of on-the-fly EUP in the context of care assistance to address caregivers' extremely variable and often hectic workflows, motivated by our past work on needs of caregivers \cite{stegner2022designing}.
Further exploration into different domains, such as by evaluating a ``Tabula 2.0'' with specific domain experts, will lead to a better understanding of when and how multi-modal on-the-fly EUP is best utilized.
}

\section{Conclusion}
This work contributes to a growing body of robot end-user programming (EUP) research by investigating end user experience with on-the-fly robot EUP. Using an open-source multimodal EUP prototype, we asked participants to create robot programs for structured and open-ended scenarios, then we interviewed them about their experiences. Our findings, which consist of five themes, illustrate user experiences which may generalize to other on-the-fly EUP tools, as well specific experiences based on Tabula's implementation. Contextualizing our findings within prior work, we develop design implications that can offer insights to inform the future design of on-the-fly EUP tools for robots.

\modified{
% Intro material
The evaluation of Tabula presented in this paper contributes to the larger cause of democratization of robots. As robots are becoming more prevalent in day-to-day life, it is even more critical for non-roboticists and non-programmers to be able to effectively utilize them. Domain experts in fields such as healthcare, education, hospitality, and service are primed to receive assistance from robots, and yet we do not yet have the intuitive, natural interfaces necessary for them to use emerging robotics tools and platforms. This work is a foundational step toward these interfaces as it is an initial exploration into a promising new EUP paradigm. 
%While this paper alone does not address all of the challenges of robot EUP, it does contribute an in-depth understanding of one specific interface and paradigm.
% Discussion material
Over time, as more novel interfaces are built and paradigms are evaluated within the research and industry communities, a pool of knowledge will accumulate to help future designers understand what kind of interface to create for a specific robot application and the user population. These tools will ultimately help more people engage with robots to create solutions that address their unique needs.
% David would delete the below sentence
%This diversification of robot end users will lead to more creative and non-conventional uses of robots, which will help us further understand how robots can be used to benefit society.
}

\begin{acks}
This work is supported by National Science Foundation (NSF) award IIS-1925043 and an NSF Graduate Research Fellowship under Grant No. DGE-1747503. DP's contributions occurred while supported as an NRC Postdoctoral Research Associate at the U.S. Naval Research Laboratory.
Any opinions, findings, and conclusions or recommendations expressed in this material are those of the authors and do not necessarily reflect the views of the NSF or the U.S. Navy.
%hailey for manuscript review
\end{acks}

\balance
\bibliographystyle{ACM-Reference-Format}
\bibliography{ref}


\begin{thebibliography}{72}

%%% the Web address.

\ifx \showCODEN    \undefined \def \showCODEN     #1{\unskip}     \fi
\ifx \showDOI      \undefined \def \showDOI       #1{#1}\fi
\ifx \showISBNx    \undefined \def \showISBNx     #1{\unskip}     \fi
\ifx \showISBNxiii \undefined \def \showISBNxiii  #1{\unskip}     \fi
\ifx \showISSN     \undefined \def \showISSN      #1{\unskip}     \fi
\ifx \showLCCN     \undefined \def \showLCCN      #1{\unskip}     \fi
\ifx \shownote     \undefined \def \shownote      #1{#1}          \fi
\ifx \showarticletitle \undefined \def \showarticletitle #1{#1}   \fi
\ifx \showURL      \undefined \def \showURL       {\relax}        \fi
% invisible to TeX
\providecommand\bibfield[2]{#2}
\providecommand\bibinfo[2]{#2}
\providecommand\natexlab[1]{#1}
\providecommand\showeprint[2][]{arXiv:#2}

\bibitem[Ajaykumar et~al\mbox{.}(2021)]%
        {ajaykumar2021survey}
\bibfield{author}{\bibinfo{person}{Gopika Ajaykumar}, \bibinfo{person}{Maureen Steele}, {and} \bibinfo{person}{Chien-Ming Huang}.} \bibinfo{year}{2021}\natexlab{}.
\newblock \showarticletitle{A survey on end-user robot programming}.
\newblock \bibinfo{journal}{\emph{ACM Computing Surveys (CSUR)}} \bibinfo{volume}{54}, \bibinfo{number}{8} (\bibinfo{year}{2021}), \bibinfo{pages}{1--36}.
\newblock
\urldef\tempurl%
\url{https://doi.org/10.1145/3466819}
\showDOI{\tempurl}


\bibitem[Alexandrova et~al\mbox{.}(2015)]%
        {alexandrova2015roboflow}
\bibfield{author}{\bibinfo{person}{Sonya Alexandrova}, \bibinfo{person}{Zachary Tatlock}, {and} \bibinfo{person}{Maya Cakmak}.} \bibinfo{year}{2015}\natexlab{}.
\newblock \showarticletitle{RoboFlow: A flow-based visual programming language for mobile manipulation tasks}. In \bibinfo{booktitle}{\emph{2015 IEEE International Conference on Robotics and Automation (ICRA)}}. IEEE, \bibinfo{pages}{5537--5544}.
\newblock
\urldef\tempurl%
\url{https://doi.org/10.1109/ICRA.2015.7139973}
\showDOI{\tempurl}


\bibitem[Alves-Oliveira et~al\mbox{.}(2022)]%
        {alves2022flex}
\bibfield{author}{\bibinfo{person}{Patricia Alves-Oliveira}, \bibinfo{person}{Kai Mihata}, \bibinfo{person}{Raida Karim}, \bibinfo{person}{Elin~A Bjorling}, {and} \bibinfo{person}{Maya Cakmak}.} \bibinfo{year}{2022}\natexlab{}.
\newblock \showarticletitle{FLEX-SDK: An Open-Source Software Development Kit for Creating Social Robots}. In \bibinfo{booktitle}{\emph{Proceedings of the 35th Annual ACM Symposium on User Interface Software and Technology}}. \bibinfo{pages}{1--10}.
\newblock
\urldef\tempurl%
\url{https://doi.org/10.1145/3526113.3545707}
\showDOI{\tempurl}


\bibitem[Amershi et~al\mbox{.}(2019)]%
        {amershi2019guidelines}
\bibfield{author}{\bibinfo{person}{Saleema Amershi}, \bibinfo{person}{Dan Weld}, \bibinfo{person}{Mihaela Vorvoreanu}, \bibinfo{person}{Adam Fourney}, \bibinfo{person}{Besmira Nushi}, \bibinfo{person}{Penny Collisson}, \bibinfo{person}{Jina Suh}, \bibinfo{person}{Shamsi Iqbal}, \bibinfo{person}{Paul~N Bennett}, \bibinfo{person}{Kori Inkpen}, {et~al\mbox{.}}} \bibinfo{year}{2019}\natexlab{}.
\newblock \showarticletitle{Guidelines for Human-AI Interaction}. In \bibinfo{booktitle}{\emph{Proceedings of the 2019 CHI Conference on Human Factors in Computing systems}}. \bibinfo{pages}{1--13}.
\newblock
\urldef\tempurl%
\url{https://doi.org/10.1145/3290605.3300233}
\showDOI{\tempurl}


\bibitem[Arawjo et~al\mbox{.}(2023)]%
        {arawjo2023chainforge}
\bibfield{author}{\bibinfo{person}{Ian Arawjo}, \bibinfo{person}{Chelse Swoopes}, \bibinfo{person}{Priyan Vaithilingam}, \bibinfo{person}{Martin Wattenberg}, {and} \bibinfo{person}{Elena Glassman}.} \bibinfo{year}{2023}\natexlab{}.
\newblock \showarticletitle{ChainForge: A Visual Toolkit for Prompt Engineering and LLM Hypothesis Testing}.
\newblock \bibinfo{journal}{\emph{arXiv preprint arXiv:2309.09128}} (\bibinfo{year}{2023}).
\newblock
\urldef\tempurl%
\url{https://doi.org/10.48550/arXiv.2309.09128}
\showDOI{\tempurl}


\bibitem[Bangor et~al\mbox{.}(2009)]%
        {bangor2009determining}
\bibfield{author}{\bibinfo{person}{Aaron Bangor}, \bibinfo{person}{Philip Kortum}, {and} \bibinfo{person}{James Miller}.} \bibinfo{year}{2009}\natexlab{}.
\newblock \showarticletitle{Determining What Individual SUS Scores Mean: Adding an Adjective Rating Scale}.
\newblock \bibinfo{journal}{\emph{J. Usability Studies}} \bibinfo{volume}{4}, \bibinfo{number}{3} (\bibinfo{date}{may} \bibinfo{year}{2009}), \bibinfo{pages}{114–123}.
\newblock
\urldef\tempurl%
\url{https://doi.org/10.5555/2835587.2835589}
\showDOI{\tempurl}


\bibitem[Barricelli et~al\mbox{.}(2019)]%
        {barricelli2019end}
\bibfield{author}{\bibinfo{person}{Barbara~Rita Barricelli}, \bibinfo{person}{Fabio Cassano}, \bibinfo{person}{Daniela Fogli}, {and} \bibinfo{person}{Antonio Piccinno}.} \bibinfo{year}{2019}\natexlab{}.
\newblock \showarticletitle{End-user development, end-user programming and end-user software engineering: A systematic mapping study}.
\newblock \bibinfo{journal}{\emph{Journal of Systems and Software}}  \bibinfo{volume}{149} (\bibinfo{year}{2019}), \bibinfo{pages}{101--137}.
\newblock
\urldef\tempurl%
\url{https://doi.org/10.1016/j.jss.2018.11.041}
\showDOI{\tempurl}


\bibitem[Beer et~al\mbox{.}(2014)]%
        {beer2014toward}
\bibfield{author}{\bibinfo{person}{Jenay~M Beer}, \bibinfo{person}{Arthur~D Fisk}, {and} \bibinfo{person}{Wendy~A Rogers}.} \bibinfo{year}{2014}\natexlab{}.
\newblock \showarticletitle{Toward a framework for levels of robot autonomy in human-robot interaction}.
\newblock \bibinfo{journal}{\emph{Journal of human-robot interaction}} \bibinfo{volume}{3}, \bibinfo{number}{2} (\bibinfo{year}{2014}), \bibinfo{pages}{74}.
\newblock
\urldef\tempurl%
\url{https://doi.org/10.5898/JHRI.3.2.Beer}
\showDOI{\tempurl}


\bibitem[Benzon(2023)]%
        {benzon2023discursive}
\bibfield{author}{\bibinfo{person}{William~L Benzon}.} \bibinfo{year}{2023}\natexlab{}.
\newblock \showarticletitle{Discursive Competence in ChatGPT, Part 1: Talking with Dragons}.
\newblock  (\bibinfo{year}{2023}).
\newblock
\urldef\tempurl%
\url{https://doi.org/10.2139/ssrn.4318832}
\showDOI{\tempurl}


\bibitem[Beschi et~al\mbox{.}(2019)]%
        {beschi2019capirci}
\bibfield{author}{\bibinfo{person}{Sara Beschi}, \bibinfo{person}{Daniela Fogli}, {and} \bibinfo{person}{Fabio Tampalini}.} \bibinfo{year}{2019}\natexlab{}.
\newblock \showarticletitle{CAPIRCI: a multi-modal system for collaborative robot programming}. In \bibinfo{booktitle}{\emph{End-User Development: 7th International Symposium, IS-EUD 2019, Hatfield, UK, July 10--12, 2019, Proceedings 7}}. Springer, \bibinfo{pages}{51--66}.
\newblock
\urldef\tempurl%
\url{https://doi.org/10.1007/978-3-030-24781-2_4}
\showDOI{\tempurl}


\bibitem[Braun and Clarke(2006)]%
        {braun2006using}
\bibfield{author}{\bibinfo{person}{Virginia Braun} {and} \bibinfo{person}{Victoria Clarke}.} \bibinfo{year}{2006}\natexlab{}.
\newblock \showarticletitle{Using thematic analysis in psychology}.
\newblock \bibinfo{journal}{\emph{Qualitative research in psychology}} \bibinfo{volume}{3}, \bibinfo{number}{2} (\bibinfo{year}{2006}), \bibinfo{pages}{77--101}.
\newblock
\urldef\tempurl%
\url{https://doi.org/10.1191/1478088706qp063oa}
\showDOI{\tempurl}


\bibitem[Brooke(1996)]%
        {brooke1996sus}
\bibfield{author}{\bibinfo{person}{John Brooke}.} \bibinfo{year}{1996}\natexlab{}.
\newblock \showarticletitle{SUS: A `Quick and Dirty' Usability Scale}.
\newblock \bibinfo{journal}{\emph{Usability Evaluation in Industry}} \bibinfo{volume}{189}, \bibinfo{number}{3} (\bibinfo{year}{1996}), \bibinfo{pages}{189--194}.
\newblock
\urldef\tempurl%
\url{https://doi.org/10.1201/9781498710411}
\showDOI{\tempurl}


\bibitem[Buchina et~al\mbox{.}(2016)]%
        {buchina2016design}
\bibfield{author}{\bibinfo{person}{Nina Buchina}, \bibinfo{person}{Sherin Kamel}, {and} \bibinfo{person}{Emilia Barakova}.} \bibinfo{year}{2016}\natexlab{}.
\newblock \showarticletitle{Design and evaluation of an end-user friendly tool for robot programming}. In \bibinfo{booktitle}{\emph{2016 25th IEEE International Symposium on Robot and Human Interactive Communication (RO-MAN)}}. IEEE, \bibinfo{pages}{185--191}.
\newblock
\urldef\tempurl%
\url{https://doi.org/10.1109/ROMAN.2016.7745109}
\showDOI{\tempurl}


\bibitem[Buchina et~al\mbox{.}(2019)]%
        {buchina2019natural}
\bibfield{author}{\bibinfo{person}{Nina~G Buchina}, \bibinfo{person}{Paula Sterkenburg}, \bibinfo{person}{Tino Lourens}, {and} \bibinfo{person}{Emilia~I Barakova}.} \bibinfo{year}{2019}\natexlab{}.
\newblock \showarticletitle{Natural language interface for programming sensory-enabled scenarios for human-robot interaction}. In \bibinfo{booktitle}{\emph{2019 28th IEEE International Conference on Robot and Human Interactive Communication (RO-MAN)}}. IEEE, \bibinfo{pages}{1--8}.
\newblock
\urldef\tempurl%
\url{https://doi.org/10.1109/RO-MAN46459.2019.8956248}
\showDOI{\tempurl}


\bibitem[Cao et~al\mbox{.}(2019a)]%
        {cao2019ghostar}
\bibfield{author}{\bibinfo{person}{Yuanzhi Cao}, \bibinfo{person}{Tianyi Wang}, \bibinfo{person}{Xun Qian}, \bibinfo{person}{Pawan~S Rao}, \bibinfo{person}{Manav Wadhawan}, \bibinfo{person}{Ke Huo}, {and} \bibinfo{person}{Karthik Ramani}.} \bibinfo{year}{2019}\natexlab{a}.
\newblock \showarticletitle{GhostAR: A time-space editor for embodied authoring of human-robot collaborative task with augmented reality}. In \bibinfo{booktitle}{\emph{Proceedings of the 32nd Annual ACM Symposium on User Interface Software and Technology}}. \bibinfo{pages}{521--534}.
\newblock
\urldef\tempurl%
\url{https://doi.org/10.1145/3332165.3347902}
\showDOI{\tempurl}


\bibitem[Cao et~al\mbox{.}(2019b)]%
        {cao2019v}
\bibfield{author}{\bibinfo{person}{Yuanzhi Cao}, \bibinfo{person}{Zhuangying Xu}, \bibinfo{person}{Fan Li}, \bibinfo{person}{Wentao Zhong}, \bibinfo{person}{Ke Huo}, {and} \bibinfo{person}{Karthik Ramani}.} \bibinfo{year}{2019}\natexlab{b}.
\newblock \showarticletitle{V. ra: An in-situ visual authoring system for robot-iot task planning with augmented reality}. In \bibinfo{booktitle}{\emph{Proceedings of the 2019 on designing interactive systems conference}}. \bibinfo{pages}{1059--1070}.
\newblock
\urldef\tempurl%
\url{https://doi.org/10.1145/3322276.3322278}
\showDOI{\tempurl}


\bibitem[Cha et~al\mbox{.}(2015)]%
        {cha2015perceived}
\bibfield{author}{\bibinfo{person}{Elizabeth Cha}, \bibinfo{person}{Anca~D Dragan}, {and} \bibinfo{person}{Siddhartha~S Srinivasa}.} \bibinfo{year}{2015}\natexlab{}.
\newblock \showarticletitle{Perceived robot capability}. In \bibinfo{booktitle}{\emph{2015 24th IEEE International Symposium on Robot and Human Interactive Communication (RO-MAN)}}. IEEE, \bibinfo{pages}{541--548}.
\newblock
\urldef\tempurl%
\url{https://doi.org/10.1109/ROMAN.2015.7333656}
\showDOI{\tempurl}


\bibitem[Chakraborti et~al\mbox{.}(2019)]%
        {chakraborti2019plan}
\bibfield{author}{\bibinfo{person}{Tathagata Chakraborti}, \bibinfo{person}{Sarath Sreedharan}, \bibinfo{person}{Sachin Grover}, {and} \bibinfo{person}{Subbarao Kambhampati}.} \bibinfo{year}{2019}\natexlab{}.
\newblock \showarticletitle{Plan explanations as model reconciliation--an empirical study}. In \bibinfo{booktitle}{\emph{2019 14th ACM/IEEE International Conference on Human-Robot Interaction (HRI)}}. IEEE, \bibinfo{pages}{258--266}.
\newblock
\urldef\tempurl%
\url{https://doi.org/10.1109/hri.2019.8673193}
\showDOI{\tempurl}


\bibitem[Chung and Cakmak(2020)]%
        {chung2020iterative}
\bibfield{author}{\bibinfo{person}{Michael Jae-Yoon Chung} {and} \bibinfo{person}{Maya Cakmak}.} \bibinfo{year}{2020}\natexlab{}.
\newblock \showarticletitle{{Iterative Repair of Social Robot Programs from Implicit User Feedback via Bayesian Inference}}. In \bibinfo{booktitle}{\emph{Proceedings of Robotics: Science and Systems}}. \bibinfo{address}{Corvalis, Oregon, USA}.
\newblock
\urldef\tempurl%
\url{https://doi.org/10.15607/RSS.2020.XVI.028}
\showDOI{\tempurl}


\bibitem[Chung et~al\mbox{.}(2016)]%
        {chung2016iterative}
\bibfield{author}{\bibinfo{person}{Michael Jae-Yoon Chung}, \bibinfo{person}{Justin Huang}, \bibinfo{person}{Leila Takayama}, \bibinfo{person}{Tessa Lau}, {and} \bibinfo{person}{Maya Cakmak}.} \bibinfo{year}{2016}\natexlab{}.
\newblock \showarticletitle{Iterative design of a system for programming socially interactive service robots}. In \bibinfo{booktitle}{\emph{Social Robotics: 8th International Conference, ICSR 2016, Kansas City, MO, USA, November 1-3, 2016 Proceedings 8}}. Springer, \bibinfo{pages}{919--929}.
\newblock
\urldef\tempurl%
\url{https://doi.org/10.1007/978-3-319-47437-3_90}
\showDOI{\tempurl}


\bibitem[Coronado et~al\mbox{.}(2021)]%
        {coronado2021towards}
\bibfield{author}{\bibinfo{person}{Enrique Coronado}, \bibinfo{person}{Dominique Deuff}, \bibinfo{person}{Pamela Carreno-Medrano}, \bibinfo{person}{Leimin Tian}, \bibinfo{person}{Dana Kuli{\'c}}, \bibinfo{person}{Shanti Sumartojo}, \bibinfo{person}{Fulvio Mastrogiovanni}, {and} \bibinfo{person}{Gentiane Venture}.} \bibinfo{year}{2021}\natexlab{}.
\newblock \showarticletitle{Towards a modular and distributed end-user development framework for human-robot interaction}.
\newblock \bibinfo{journal}{\emph{IEEE Access}}  \bibinfo{volume}{9} (\bibinfo{year}{2021}), \bibinfo{pages}{12675--12692}.
\newblock
\urldef\tempurl%
\url{https://doi.org/10.1109/ACCESS.2021.3051605}
\showDOI{\tempurl}


\bibitem[Correa et~al\mbox{.}(2010)]%
        {correa2010multimodal}
\bibfield{author}{\bibinfo{person}{Andrew Correa}, \bibinfo{person}{Matthew~R Walter}, \bibinfo{person}{Luke Fletcher}, \bibinfo{person}{Jim Glass}, \bibinfo{person}{Seth Teller}, {and} \bibinfo{person}{Randall Davis}.} \bibinfo{year}{2010}\natexlab{}.
\newblock \showarticletitle{Multimodal interaction with an autonomous forklift}. In \bibinfo{booktitle}{\emph{2010 5th ACM/IEEE International Conference on Human-Robot Interaction (HRI)}}. IEEE, \bibinfo{pages}{243--250}.
\newblock
\urldef\tempurl%
\url{https://doi.org/10.1109/HRI.2010.5453188}
\showDOI{\tempurl}


\bibitem[De~Russis and Corno(2015)]%
        {de2015homerules}
\bibfield{author}{\bibinfo{person}{Luigi De~Russis} {and} \bibinfo{person}{Fulvio Corno}.} \bibinfo{year}{2015}\natexlab{}.
\newblock \showarticletitle{Homerules: A tangible end-user programming interface for smart homes}. In \bibinfo{booktitle}{\emph{Proceedings of the 33rd Annual ACM Conference Extended Abstracts on Human Factors in Computing Systems}}. \bibinfo{pages}{2109--2114}.
\newblock
\urldef\tempurl%
\url{https://doi.org/10.1145/2702613.2732795}
\showDOI{\tempurl}


\bibitem[Forbes et~al\mbox{.}(2015)]%
        {forbes2015robot}
\bibfield{author}{\bibinfo{person}{Maxwell Forbes}, \bibinfo{person}{Rajesh~PN Rao}, \bibinfo{person}{Luke Zettlemoyer}, {and} \bibinfo{person}{Maya Cakmak}.} \bibinfo{year}{2015}\natexlab{}.
\newblock \showarticletitle{Robot programming by demonstration with situated spatial language understanding}. In \bibinfo{booktitle}{\emph{2015 IEEE International Conference on Robotics and Automation (ICRA)}}. IEEE, \bibinfo{pages}{2014--2020}.
\newblock
\urldef\tempurl%
\url{https://doi.org/10.1109/ICRA.2015.7139462}
\showDOI{\tempurl}


\bibitem[Friard and Gamba(2016)]%
        {friard2016boris}
\bibfield{author}{\bibinfo{person}{Olivier Friard} {and} \bibinfo{person}{Marco Gamba}.} \bibinfo{year}{2016}\natexlab{}.
\newblock \showarticletitle{BORIS: a free, versatile open-source event-logging software for video/audio coding and live observations}.
\newblock \bibinfo{journal}{\emph{Methods in ecology and evolution}} \bibinfo{volume}{7}, \bibinfo{number}{11} (\bibinfo{year}{2016}), \bibinfo{pages}{1325--1330}.
\newblock
\urldef\tempurl%
\url{https://doi.org/10.1111/2041-210X.12584}
\showDOI{\tempurl}


\bibitem[Gao and Huang(2019)]%
        {gao2019pati}
\bibfield{author}{\bibinfo{person}{Yuxiang Gao} {and} \bibinfo{person}{Chien-Ming Huang}.} \bibinfo{year}{2019}\natexlab{}.
\newblock \showarticletitle{PATI: a projection-based augmented table-top interface for robot programming}. In \bibinfo{booktitle}{\emph{Proceedings of the 24th international conference on intelligent user interfaces}}. \bibinfo{pages}{345--355}.
\newblock
\urldef\tempurl%
\url{https://doi.org/10.1145/3301275.3302326}
\showDOI{\tempurl}


\bibitem[Glas et~al\mbox{.}(2011)]%
        {glas2012interaction}
\bibfield{author}{\bibinfo{person}{Dylan Glas}, \bibinfo{person}{Satoru Satake}, \bibinfo{person}{Takayuki Kanda}, {and} \bibinfo{person}{Norihiro Hagita}.} \bibinfo{year}{2011}\natexlab{}.
\newblock \showarticletitle{An Interaction Design Framework for Social Robots}. In \bibinfo{booktitle}{\emph{Proceedings of Robotics: Science and Systems}}. \bibinfo{address}{Los Angeles, CA, USA}.
\newblock
\urldef\tempurl%
\url{https://doi.org/10.15607/RSS.2011.VII.014}
\showDOI{\tempurl}


\bibitem[Gomaa and Scott(1981)]%
        {gomaa1981prototyping}
\bibfield{author}{\bibinfo{person}{Hassan Gomaa} {and} \bibinfo{person}{Douglas~BH Scott}.} \bibinfo{year}{1981}\natexlab{}.
\newblock \showarticletitle{Prototyping as a tool in the specification of user requirements}. In \bibinfo{booktitle}{\emph{Proceedings of the 5th international conference on Software engineering}}. \bibinfo{pages}{333--342}.
\newblock
\urldef\tempurl%
\url{https://dl.acm.org/doi/10.5555/800078.802546}
\showURL{%
\tempurl}


\bibitem[Good(1999)]%
        {good1999vpls}
\bibfield{author}{\bibinfo{person}{Judith Good}.} \bibinfo{year}{1999}\natexlab{}.
\newblock \showarticletitle{VPLs and novice program comprehension: How do different languages compare?}. In \bibinfo{booktitle}{\emph{Proceedings 1999 IEEE Symposium on Visual Languages}}. IEEE, \bibinfo{pages}{262--269}.
\newblock
\urldef\tempurl%
\url{https://doi.org/10.1109/VL.1999.795912}
\showDOI{\tempurl}


\bibitem[Gorostiza and Salichs(2011)]%
        {gorostiza2011end}
\bibfield{author}{\bibinfo{person}{Javi~F Gorostiza} {and} \bibinfo{person}{Miguel~A Salichs}.} \bibinfo{year}{2011}\natexlab{}.
\newblock \showarticletitle{End-user programming of a social robot by dialog}.
\newblock \bibinfo{journal}{\emph{Robotics and Autonomous Systems}} \bibinfo{volume}{59}, \bibinfo{number}{12} (\bibinfo{year}{2011}), \bibinfo{pages}{1102--1114}.
\newblock
\urldef\tempurl%
\url{https://doi.org/10.1016/j.robot.2011.07.009}
\showDOI{\tempurl}


\bibitem[Green and Petre(1996)]%
        {green1996usability}
\bibfield{author}{\bibinfo{person}{Thomas R.~G. Green} {and} \bibinfo{person}{Marian Petre}.} \bibinfo{year}{1996}\natexlab{}.
\newblock \showarticletitle{Usability analysis of visual programming environments: a ‘cognitive dimensions’ framework}.
\newblock \bibinfo{journal}{\emph{Journal of Visual Languages \& Computing}} \bibinfo{volume}{7}, \bibinfo{number}{2} (\bibinfo{year}{1996}), \bibinfo{pages}{131--174}.
\newblock
\urldef\tempurl%
\url{https://doi.org/10.1006/jvlc.1996.0009}
\showDOI{\tempurl}


\bibitem[Huang et~al\mbox{.}(2020)]%
        {huang2020vipo}
\bibfield{author}{\bibinfo{person}{Gaoping Huang}, \bibinfo{person}{Pawan~S Rao}, \bibinfo{person}{Meng-Han Wu}, \bibinfo{person}{Xun Qian}, \bibinfo{person}{Shimon~Y Nof}, \bibinfo{person}{Karthik Ramani}, {and} \bibinfo{person}{Alexander~J Quinn}.} \bibinfo{year}{2020}\natexlab{}.
\newblock \showarticletitle{Vipo: Spatial-visual programming with functions for robot-IoT workflows}. In \bibinfo{booktitle}{\emph{Proceedings of the 2020 CHI Conference on Human Factors in Computing Systems}}. \bibinfo{pages}{1--13}.
\newblock
\urldef\tempurl%
\url{https://doi.org/10.1145/3313831.3376670}
\showDOI{\tempurl}


\bibitem[Huang and Cakmak(2015)]%
        {huang2016trigger}
\bibfield{author}{\bibinfo{person}{Justin Huang} {and} \bibinfo{person}{Maya Cakmak}.} \bibinfo{year}{2015}\natexlab{}.
\newblock \showarticletitle{Supporting mental model accuracy in trigger-action programming}. In \bibinfo{booktitle}{\emph{Proceedings of the 2015 ACM International Joint Conference on Pervasive and Ubiquitous Computing}} (Osaka, Japan) \emph{(\bibinfo{series}{UbiComp '15})}. \bibinfo{publisher}{Association for Computing Machinery}, \bibinfo{address}{New York, NY, USA}, \bibinfo{pages}{215–225}.
\newblock
\showISBNx{9781450335744}
\urldef\tempurl%
\url{https://doi.org/10.1145/2750858.2805830}
\showDOI{\tempurl}


\bibitem[Huang and Cakmak(2017)]%
        {huang2017code3}
\bibfield{author}{\bibinfo{person}{Justin Huang} {and} \bibinfo{person}{Maya Cakmak}.} \bibinfo{year}{2017}\natexlab{}.
\newblock \showarticletitle{Code3: A system for end-to-end programming of mobile manipulator robots for novices and experts}. In \bibinfo{booktitle}{\emph{Proceedings of the 2017 ACM/IEEE International Conference on Human-Robot Interaction}}. \bibinfo{pages}{453--462}.
\newblock
\urldef\tempurl%
\url{https://doi.org/10.1145/2909824.3020215}
\showDOI{\tempurl}


\bibitem[Jiang et~al\mbox{.}(2022)]%
        {jiang2022promptmaker}
\bibfield{author}{\bibinfo{person}{Ellen Jiang}, \bibinfo{person}{Kristen Olson}, \bibinfo{person}{Edwin Toh}, \bibinfo{person}{Alejandra Molina}, \bibinfo{person}{Aaron Donsbach}, \bibinfo{person}{Michael Terry}, {and} \bibinfo{person}{Carrie~J Cai}.} \bibinfo{year}{2022}\natexlab{}.
\newblock \showarticletitle{Promptmaker: Prompt-based prototyping with large language models}. In \bibinfo{booktitle}{\emph{CHI Conference on Human Factors in Computing Systems Extended Abstracts}}. \bibinfo{pages}{1--8}.
\newblock
\urldef\tempurl%
\url{https://doi.org/10.1145/3491101.3503564}
\showDOI{\tempurl}


\bibitem[Kambhamettu et~al\mbox{.}(2021)]%
        {kambhamettu2021collecting}
\bibfield{author}{\bibinfo{person}{Rajeswari~Hita Kambhamettu}, \bibinfo{person}{Michael Jae-Yoon~Chung}, \bibinfo{person}{Vinitha Ranganeni}, {and} \bibinfo{person}{Patr{\'\i}cia Alves-Oliveira}.} \bibinfo{year}{2021}\natexlab{}.
\newblock \showarticletitle{Collecting Insights about How Novice Programmers Naturally Express Programs for Robots}. Plateau Workshop.
\newblock
\urldef\tempurl%
\url{https://doi.org/10.1184/R1/19799197.v1}
\showDOI{\tempurl}


\bibitem[Kassel and Malloy(2003)]%
        {kassel2003approach}
\bibfield{author}{\bibinfo{person}{Neil~W Kassel} {and} \bibinfo{person}{Brian~A Malloy}.} \bibinfo{year}{2003}\natexlab{}.
\newblock \showarticletitle{An approach to automate requirements elicitation and specification}. In \bibinfo{booktitle}{\emph{Proc. of the 7th Int. Conf. on Software Engineering and Applications}}. Citeseer, \bibinfo{pages}{3--5}.
\newblock


\bibitem[Ko et~al\mbox{.}(2011)]%
        {ko2011state}
\bibfield{author}{\bibinfo{person}{Amy~J Ko}, \bibinfo{person}{Robin Abraham}, \bibinfo{person}{Laura Beckwith}, \bibinfo{person}{Alan Blackwell}, \bibinfo{person}{Margaret Burnett}, \bibinfo{person}{Martin Erwig}, \bibinfo{person}{Chris Scaffidi}, \bibinfo{person}{Joseph Lawrance}, \bibinfo{person}{Henry Lieberman}, \bibinfo{person}{Brad Myers}, {et~al\mbox{.}}} \bibinfo{year}{2011}\natexlab{}.
\newblock \showarticletitle{The state of the art in end-user software engineering}.
\newblock \bibinfo{journal}{\emph{ACM Computing Surveys (CSUR)}} \bibinfo{volume}{43}, \bibinfo{number}{3} (\bibinfo{year}{2011}), \bibinfo{pages}{1--44}.
\newblock
\urldef\tempurl%
\url{https://doi.org/10.1145/1922649.1922658}
\showDOI{\tempurl}


\bibitem[Landis and Koch(1977)]%
        {landis1977measurement}
\bibfield{author}{\bibinfo{person}{J~Richard Landis} {and} \bibinfo{person}{Gary~G Koch}.} \bibinfo{year}{1977}\natexlab{}.
\newblock \showarticletitle{The measurement of observer agreement for categorical data}.
\newblock \bibinfo{journal}{\emph{biometrics}} (\bibinfo{year}{1977}), \bibinfo{pages}{159--174}.
\newblock
\urldef\tempurl%
\url{https://doi.org/10.2307/2529310}
\showDOI{\tempurl}


\bibitem[Leonardi et~al\mbox{.}(2019)]%
        {leonardi2019trigger}
\bibfield{author}{\bibinfo{person}{Nicola Leonardi}, \bibinfo{person}{Marco Manca}, \bibinfo{person}{Fabio Patern{\`o}}, {and} \bibinfo{person}{Carmen Santoro}.} \bibinfo{year}{2019}\natexlab{}.
\newblock \showarticletitle{Trigger-action programming for personalising humanoid robot behaviour}. In \bibinfo{booktitle}{\emph{Proceedings of the 2019 CHI Conference on Human Factors in Computing Systems}}. \bibinfo{pages}{1--13}.
\newblock
\urldef\tempurl%
\url{https://doi.org/10.1145/3290605.3300675}
\showDOI{\tempurl}


\bibitem[Li et~al\mbox{.}(2019)]%
        {li2019pumice}
\bibfield{author}{\bibinfo{person}{Toby Jia-Jun Li}, \bibinfo{person}{Marissa Radensky}, \bibinfo{person}{Justin Jia}, \bibinfo{person}{Kirielle Singarajah}, \bibinfo{person}{Tom~M. Mitchell}, {and} \bibinfo{person}{Brad~A. Myers}.} \bibinfo{year}{2019}\natexlab{}.
\newblock \showarticletitle{PUMICE: A Multi-Modal Agent that Learns Concepts and Conditionals from Natural Language and Demonstrations}. In \bibinfo{booktitle}{\emph{Proceedings of the 32nd Annual ACM Symposium on User Interface Software and Technology}} (New Orleans, LA, USA) \emph{(\bibinfo{series}{UIST '19})}. \bibinfo{publisher}{Association for Computing Machinery}, \bibinfo{address}{New York, NY, USA}, \bibinfo{pages}{577–589}.
\newblock
\showISBNx{9781450368162}
\urldef\tempurl%
\url{https://doi.org/10.1145/3332165.3347899}
\showDOI{\tempurl}


\bibitem[Lieberman et~al\mbox{.}(2006)]%
        {lieberman2006end}
\bibfield{author}{\bibinfo{person}{Henry Lieberman}, \bibinfo{person}{Fabio Patern{\`o}}, \bibinfo{person}{Markus Klann}, {and} \bibinfo{person}{Volker Wulf}.} \bibinfo{year}{2006}\natexlab{}.
\newblock \showarticletitle{End-user development: An emerging paradigm}.
\newblock In \bibinfo{booktitle}{\emph{End user development}}. \bibinfo{publisher}{Springer}, \bibinfo{pages}{1--8}.
\newblock
\urldef\tempurl%
\url{https://doi.org/10.1007/1-4020-5386-X_1}
\showDOI{\tempurl}


\bibitem[Little et~al\mbox{.}(2010)]%
        {little2010sloppy}
\bibfield{author}{\bibinfo{person}{Greg Little}, \bibinfo{person}{Robert~C Miller}, \bibinfo{person}{Victoria~H Chou}, \bibinfo{person}{Michael Bernstein}, \bibinfo{person}{Tessa Lau}, {and} \bibinfo{person}{Allen Cypher}.} \bibinfo{year}{2010}\natexlab{}.
\newblock \showarticletitle{Sloppy programming}.
\newblock In \bibinfo{booktitle}{\emph{No Code Required}}. \bibinfo{publisher}{Elsevier}, \bibinfo{pages}{289--307}.
\newblock
\urldef\tempurl%
\url{https://doi.org/10.1016/B978-0-12-381541-5.00015-8}
\showDOI{\tempurl}


\bibitem[Liu et~al\mbox{.}(2011)]%
        {liu2011roboshop}
\bibfield{author}{\bibinfo{person}{Kexi Liu}, \bibinfo{person}{Daisuke Sakamoto}, \bibinfo{person}{Masahiko Inami}, {and} \bibinfo{person}{Takeo Igarashi}.} \bibinfo{year}{2011}\natexlab{}.
\newblock \showarticletitle{Roboshop: multi-layered sketching interface for robot housework assignment and management}. In \bibinfo{booktitle}{\emph{Proceedings of the SIGCHI Conference on Human Factors in Computing Systems}}. \bibinfo{pages}{647--656}.
\newblock
\urldef\tempurl%
\url{https://doi.org/10.1145/1978942.1979035}
\showDOI{\tempurl}


\bibitem[Lu et~al\mbox{.}(2012)]%
        {lu2012structure}
\bibfield{author}{\bibinfo{person}{Fei Lu}, \bibinfo{person}{Xinran Wang}, {and} \bibinfo{person}{Guohui Tian}.} \bibinfo{year}{2012}\natexlab{}.
\newblock \showarticletitle{The structure and application of intelligent space system oriented to home service robot}. In \bibinfo{booktitle}{\emph{2012 IEEE International Conference on Information and Automation}}. \bibinfo{pages}{289--294}.
\newblock
\urldef\tempurl%
\url{https://doi.org/10.1109/ICInfA.2012.6246820}
\showDOI{\tempurl}


\bibitem[Lucci and Patern{\`o}(2014)]%
        {lucci2014understanding}
\bibfield{author}{\bibinfo{person}{Gabriella Lucci} {and} \bibinfo{person}{Fabio Patern{\`o}}.} \bibinfo{year}{2014}\natexlab{}.
\newblock \showarticletitle{Understanding end-user development of context-dependent applications in smartphones}. In \bibinfo{booktitle}{\emph{Human-Centered Software Engineering: 5th IFIP WG 13.2 International Conference, HCSE 2014, Paderborn, Germany, September 16-18, 2014. Proceedings 5}}. Springer, \bibinfo{pages}{182--198}.
\newblock
\urldef\tempurl%
\url{https://doi.org/10.1007/978-3-662-44811-3_11}
\showDOI{\tempurl}


\bibitem[MacLaurin(2009)]%
        {maclaurin2009kodu}
\bibfield{author}{\bibinfo{person}{Matt MacLaurin}.} \bibinfo{year}{2009}\natexlab{}.
\newblock \showarticletitle{Kodu: end-user programming and design for games}. In \bibinfo{booktitle}{\emph{Proceedings of the 4th international conference on foundations of digital games}}. \bibinfo{pages}{xviii--xix}.
\newblock
\urldef\tempurl%
\url{https://doi.org/10.1145/1536513.1536516}
\showDOI{\tempurl}


\bibitem[OpenAI(2023)]%
        {openai2023gpt4}
\bibfield{author}{\bibinfo{person}{OpenAI}.} \bibinfo{year}{2023}\natexlab{}.
\newblock \bibinfo{title}{GPT-4 Technical Report}.
\newblock
\newblock
\urldef\tempurl%
\url{https://doi.org/10.48550/arXiv.2303.08774}
\showDOI{\tempurl}
\showeprint[arxiv]{2303.08774}~[cs.CL]


\bibitem[Ostyakova et~al\mbox{.}(2023)]%
        {ostyakova2023linguistic}
\bibfield{author}{\bibinfo{person}{Lidiia Ostyakova}, \bibinfo{person}{Kseniia PetukhovaO}, \bibinfo{person}{Veronika Smilga}, {and} \bibinfo{person}{Dilyara ZharikovaO}.} \bibinfo{year}{2023}\natexlab{}.
\newblock \showarticletitle{Linguistic Annotation Generation with ChatGPT: a Synthetic Dataset of Speech Functions for Discourse Annotation of Casual Conversations}. In \bibinfo{booktitle}{\emph{Proceedings of the International Conference “Dialogue}}, Vol.~\bibinfo{volume}{2023}.
\newblock
\urldef\tempurl%
\url{https://doi.org/10.28995/2075-7182-2023-22-386-403}
\showDOI{\tempurl}


\bibitem[Porfirio et~al\mbox{.}(2019)]%
        {porfirio2019bodystorming}
\bibfield{author}{\bibinfo{person}{David Porfirio}, \bibinfo{person}{Evan Fisher}, \bibinfo{person}{Allison Saupp{\'e}}, \bibinfo{person}{Aws Albarghouthi}, {and} \bibinfo{person}{Bilge Mutlu}.} \bibinfo{year}{2019}\natexlab{}.
\newblock \showarticletitle{Bodystorming human-robot interactions}. In \bibinfo{booktitle}{\emph{proceedings of the 32nd annual ACM symposium on user Interface software and technology}}. \bibinfo{pages}{479--491}.
\newblock
\urldef\tempurl%
\url{https://doi.org/10.1145/3332165.3347957}
\showDOI{\tempurl}


\bibitem[Porfirio et~al\mbox{.}(2024)]%
        {porfirio2024polaris}
\bibfield{author}{\bibinfo{person}{David Porfirio}, \bibinfo{person}{Mark Roberts}, {and} \bibinfo{person}{Laura~M. Hiatt}.} \bibinfo{year}{2024}\natexlab{}.
\newblock \showarticletitle{Goal-Oriented End-User Programming of Robots}. In \bibinfo{booktitle}{\emph{Proceedings of the 2024 ACM/IEEE International Conference on Human-Robot Interaction}} (Boulder, CO, USA) \emph{(\bibinfo{series}{HRI '24})}. \bibinfo{publisher}{Association for Computing Machinery}, \bibinfo{address}{New York, NY, USA}, \bibinfo{pages}{582–591}.
\newblock
\showISBNx{9798400703225}
\urldef\tempurl%
\url{https://doi.org/10.1145/3610977.3634974}
\showDOI{\tempurl}


\bibitem[Porfirio et~al\mbox{.}(2023)]%
        {porfirio2023sketching}
\bibfield{author}{\bibinfo{person}{David Porfirio}, \bibinfo{person}{Laura Stegner}, \bibinfo{person}{Maya Cakmak}, \bibinfo{person}{Allison Saupp{\'e}}, \bibinfo{person}{Aws Albarghouthi}, {and} \bibinfo{person}{Bilge Mutlu}.} \bibinfo{year}{2023}\natexlab{}.
\newblock \showarticletitle{Sketching Robot Programs On the Fly}. In \bibinfo{booktitle}{\emph{Proceedings of the 2023 ACM/IEEE International Conference on Human-Robot Interaction}}. \bibinfo{pages}{584--593}.
\newblock
\urldef\tempurl%
\url{https://doi.org/10.1145/3568162.3576991}
\showDOI{\tempurl}


\bibitem[Porfirio et~al\mbox{.}(2021)]%
        {porfirio2021figaro}
\bibfield{author}{\bibinfo{person}{David~J Porfirio}, \bibinfo{person}{Laura Stegner}, \bibinfo{person}{Maya Cakmak}, \bibinfo{person}{Allison Saupp{\'e}}, \bibinfo{person}{Aws Albarghouthi}, {and} \bibinfo{person}{Bilge Mutlu}.} \bibinfo{year}{2021}\natexlab{}.
\newblock \showarticletitle{Figaro: A tabletop authoring environment for human-robot interaction}. In \bibinfo{booktitle}{\emph{Proceedings of the 2021 CHI Conference on Human Factors in Computing Systems}}. \bibinfo{pages}{1--15}.
\newblock
\urldef\tempurl%
\url{https://doi.org/10.1145/3411764.3446864}
\showDOI{\tempurl}


\bibitem[Pot et~al\mbox{.}(2009)]%
        {pot2009choregraphe}
\bibfield{author}{\bibinfo{person}{Emmanuel Pot}, \bibinfo{person}{J{\'e}r{\^o}me Monceaux}, \bibinfo{person}{Rodolphe Gelin}, {and} \bibinfo{person}{Bruno Maisonnier}.} \bibinfo{year}{2009}\natexlab{}.
\newblock \showarticletitle{Choregraphe: a graphical tool for humanoid robot programming}. In \bibinfo{booktitle}{\emph{RO-MAN 2009-The 18th IEEE International Symposium on Robot and Human Interactive Communication}}. IEEE, \bibinfo{pages}{46--51}.
\newblock
\urldef\tempurl%
\url{https://doi.org/10.1109/ROMAN.2009.5326209}
\showDOI{\tempurl}


\bibitem[Puig et~al\mbox{.}(2018)]%
        {puig2018virtualhome}
\bibfield{author}{\bibinfo{person}{Xavier Puig}, \bibinfo{person}{Kevin Ra}, \bibinfo{person}{Marko Boben}, \bibinfo{person}{Jiaman Li}, \bibinfo{person}{Tingwu Wang}, \bibinfo{person}{Sanja Fidler}, {and} \bibinfo{person}{Antonio Torralba}.} \bibinfo{year}{2018}\natexlab{}.
\newblock \showarticletitle{Virtualhome: Simulating household activities via programs}. In \bibinfo{booktitle}{\emph{Proceedings of the IEEE Conference on Computer Vision and Pattern Recognition}}. \bibinfo{pages}{8494--8502}.
\newblock
\urldef\tempurl%
\url{https://doi.org/10.48550/arXiv.1806.07011}
\showDOI{\tempurl}


\bibitem[Ranganeni et~al\mbox{.}(2024)]%
        {ranganeni2024robots}
\bibfield{author}{\bibinfo{person}{Vinitha Ranganeni}, \bibinfo{person}{Vy Nguyen}, \bibinfo{person}{Henry Evans}, \bibinfo{person}{Jane Evans}, \bibinfo{person}{Julian Mehu}, \bibinfo{person}{Samuel Olatunji}, \bibinfo{person}{Wendy Rogers}, \bibinfo{person}{Aaron Edsinger}, \bibinfo{person}{Charles Kemp}, {and} \bibinfo{person}{Maya Cakmak}.} \bibinfo{year}{2024}\natexlab{}.
\newblock \showarticletitle{Robots for Humanity: In-Home Deployment of Stretch RE2}. In \bibinfo{booktitle}{\emph{Companion of the 2024 ACM/IEEE International Conference on Human-Robot Interaction}}. \bibinfo{pages}{1299--1301}.
\newblock
\urldef\tempurl%
\url{https://doi.org/10.1145/3610978.3641114}
\showDOI{\tempurl}


\bibitem[Sakamoto et~al\mbox{.}(2009)]%
        {sakamoto2009sketch}
\bibfield{author}{\bibinfo{person}{Daisuke Sakamoto}, \bibinfo{person}{Koichiro Honda}, \bibinfo{person}{Masahiko Inami}, {and} \bibinfo{person}{Takeo Igarashi}.} \bibinfo{year}{2009}\natexlab{}.
\newblock \showarticletitle{Sketch and run: a stroke-based interface for home robots}. In \bibinfo{booktitle}{\emph{Proceedings of the SIGCHI conference on human factors in computing systems}}. \bibinfo{pages}{197--200}.
\newblock
\urldef\tempurl%
\url{https://doi.org/10.1145/1518701.1518733}
\showDOI{\tempurl}


\bibitem[Schoen et~al\mbox{.}(2022)]%
        {schoen2022coframe}
\bibfield{author}{\bibinfo{person}{Andrew Schoen}, \bibinfo{person}{Nathan White}, \bibinfo{person}{Curt Henrichs}, \bibinfo{person}{Amanda Siebert-Evenstone}, \bibinfo{person}{David Shaffer}, {and} \bibinfo{person}{Bilge Mutlu}.} \bibinfo{year}{2022}\natexlab{}.
\newblock \showarticletitle{CoFrame: A System for Training Novice Cabot Programmers}. In \bibinfo{booktitle}{\emph{2022 17th ACM/IEEE International Conference on Human-Robot Interaction (HRI)}}. IEEE, \bibinfo{pages}{185--194}.
\newblock
\urldef\tempurl%
\url{https://doi.org/10.1109/HRI53351.2022.9889345}
\showDOI{\tempurl}


\bibitem[Senft et~al\mbox{.}(2021)]%
        {senft2021task}
\bibfield{author}{\bibinfo{person}{Emmanuel Senft}, \bibinfo{person}{Michael Hagenow}, \bibinfo{person}{Kevin Welsh}, \bibinfo{person}{Robert Radwin}, \bibinfo{person}{Michael Zinn}, \bibinfo{person}{Michael Gleicher}, {and} \bibinfo{person}{Bilge Mutlu}.} \bibinfo{year}{2021}\natexlab{}.
\newblock \showarticletitle{Task-level authoring for remote robot teleoperation}.
\newblock \bibinfo{journal}{\emph{Frontiers in Robotics and AI}}  \bibinfo{volume}{8} (\bibinfo{year}{2021}), \bibinfo{pages}{707149}.
\newblock
\urldef\tempurl%
\url{https://doi.org/10.3389/frobt.2021.707149}
\showDOI{\tempurl}


\bibitem[Senft et~al\mbox{.}(2020)]%
        {senft2020would}
\bibfield{author}{\bibinfo{person}{Emmanuel Senft}, \bibinfo{person}{Satoru Satake}, {and} \bibinfo{person}{Takayuki Kanda}.} \bibinfo{year}{2020}\natexlab{}.
\newblock \showarticletitle{Would You Mind Me if I Pass by You? Socially-Appropriate Behaviour for an Omni-based Social Robot in Narrow Environment}. In \bibinfo{booktitle}{\emph{Proceedings of the 2020 ACM/IEEE International Conference on Human-Robot Interaction}}. \bibinfo{pages}{539--547}.
\newblock
\urldef\tempurl%
\url{https://doi.org/10.1145/3319502.3374812}
\showDOI{\tempurl}


\bibitem[Smith(2012)]%
        {smith2012planning}
\bibfield{author}{\bibinfo{person}{David Smith}.} \bibinfo{year}{2012}\natexlab{}.
\newblock \showarticletitle{Planning as an iterative process}. In \bibinfo{booktitle}{\emph{Proceedings of the AAAI Conference on Artificial Intelligence}}, Vol.~\bibinfo{volume}{26}. \bibinfo{pages}{2180--2185}.
\newblock
\urldef\tempurl%
\url{https://doi.org/10.1609/aaai.v26i1.8449}
\showDOI{\tempurl}


\bibitem[Stegner and Mutlu(2022)]%
        {stegner2022designing}
\bibfield{author}{\bibinfo{person}{Laura Stegner} {and} \bibinfo{person}{Bilge Mutlu}.} \bibinfo{year}{2022}\natexlab{}.
\newblock \showarticletitle{Designing for Caregiving: Integrating Robotic Assistance in Senior Living Communities}. In \bibinfo{booktitle}{\emph{Designing Interactive Systems Conference}} (Virtual Event, Australia) \emph{(\bibinfo{series}{DIS '22})}. \bibinfo{publisher}{Association for Computing Machinery}, \bibinfo{address}{New York, NY, USA}, \bibinfo{pages}{1934–1947}.
\newblock
\showISBNx{9781450393584}
\urldef\tempurl%
\url{https://doi.org/10.1145/3532106.3533536}
\showDOI{\tempurl}


\bibitem[Stegner et~al\mbox{.}(2023)]%
        {stegner2023situated}
\bibfield{author}{\bibinfo{person}{Laura Stegner}, \bibinfo{person}{Emmanuel Senft}, {and} \bibinfo{person}{Bilge Mutlu}.} \bibinfo{year}{2023}\natexlab{}.
\newblock \showarticletitle{Situated participatory design: A method for in situ design of robotic interaction with older adults}. In \bibinfo{booktitle}{\emph{Proceedings of the 2023 CHI Conference on Human Factors in Computing Systems}}. \bibinfo{pages}{1--15}.
\newblock
\urldef\tempurl%
\url{https://doi.org/10.1145/3544548.3580893}
\showDOI{\tempurl}


\bibitem[Stenmark et~al\mbox{.}(2017)]%
        {stenmark2017simplified}
\bibfield{author}{\bibinfo{person}{Maj Stenmark}, \bibinfo{person}{Mathias Haage}, {and} \bibinfo{person}{Elin~Anna Topp}.} \bibinfo{year}{2017}\natexlab{}.
\newblock \showarticletitle{Simplified programming of re-usable skills on a safe industrial robot: Prototype and evaluation}. In \bibinfo{booktitle}{\emph{Proceedings of the 2017 ACM/IEEE International Conference on Human-Robot Interaction}}. \bibinfo{pages}{463--472}.
\newblock
\urldef\tempurl%
\url{https://doi.org/10.1145/2909824.3020227}
\showDOI{\tempurl}


\bibitem[Takayama(2012)]%
        {takayama2012perspectives}
\bibfield{author}{\bibinfo{person}{Leila Takayama}.} \bibinfo{year}{2012}\natexlab{}.
\newblock \showarticletitle{Perspectives on agency interacting with and through personal robots}.
\newblock In \bibinfo{booktitle}{\emph{Human-computer interaction: the agency perspective}}. \bibinfo{publisher}{Springer}, \bibinfo{pages}{195--214}.
\newblock
\urldef\tempurl%
\url{https://doi.org/10.1007/978-3-642-25691-2_8}
\showDOI{\tempurl}


\bibitem[Teller et~al\mbox{.}(2010)]%
        {teller2010voice}
\bibfield{author}{\bibinfo{person}{Seth Teller}, \bibinfo{person}{Matthew~R Walter}, \bibinfo{person}{Matthew Antone}, \bibinfo{person}{Andrew Correa}, \bibinfo{person}{Randall Davis}, \bibinfo{person}{Luke Fletcher}, \bibinfo{person}{Emilio Frazzoli}, \bibinfo{person}{Jim Glass}, \bibinfo{person}{Jonathan~P How}, \bibinfo{person}{Albert~S Huang}, {et~al\mbox{.}}} \bibinfo{year}{2010}\natexlab{}.
\newblock \showarticletitle{A voice-commandable robotic forklift working alongside humans in minimally-prepared outdoor environments}. In \bibinfo{booktitle}{\emph{2010 IEEE International Conference on Robotics and Automation}}. IEEE, \bibinfo{pages}{526--533}.
\newblock
\urldef\tempurl%
\url{https://doi.org/10.1109/ROBOT.2010.5509238}
\showDOI{\tempurl}


\bibitem[Trafton and Reiser(1991)]%
        {trafton1991providing}
\bibfield{author}{\bibinfo{person}{John~Gregory Trafton} {and} \bibinfo{person}{Brian~J Reiser}.} \bibinfo{year}{1991}\natexlab{}.
\newblock \showarticletitle{Providing natural representations to facilitate novices' understanding in a new domain: Forward and backward reasoning in programming}. In \bibinfo{booktitle}{\emph{Proceedings of the 13th Annual Conference of the Cognitive Science Society}}. \bibinfo{publisher}{Lawrence Erlbaum Associates, Inc.}, \bibinfo{pages}{923--927}.
\newblock


\bibitem[Van~Buren and Cook(1998)]%
        {van1998experiences}
\bibfield{author}{\bibinfo{person}{Jim Van~Buren} {and} \bibinfo{person}{David Cook}.} \bibinfo{year}{1998}\natexlab{}.
\newblock \showarticletitle{Experiences in the adoption of requirements engineering technologies}.
\newblock \bibinfo{journal}{\emph{Crosstalk-The Journal of Defense Software Engineering}} \bibinfo{volume}{11}, \bibinfo{number}{12} (\bibinfo{year}{1998}), \bibinfo{pages}{3--10}.
\newblock


\bibitem[Walker et~al\mbox{.}(2019)]%
        {walker2019neural}
\bibfield{author}{\bibinfo{person}{Nick Walker}, \bibinfo{person}{Yu-Tang Peng}, {and} \bibinfo{person}{Maya Cakmak}.} \bibinfo{year}{2019}\natexlab{}.
\newblock \showarticletitle{Neural semantic parsing with anonymization for command understanding in general-purpose service robots}.
\newblock In \bibinfo{booktitle}{\emph{Robot World Cup}}. \bibinfo{publisher}{Springer}, \bibinfo{pages}{337--350}.
\newblock
\urldef\tempurl%
\url{https://doi.org/10.1007/978-3-030-35699-6_26}
\showDOI{\tempurl}


\bibitem[Wang et~al\mbox{.}(2023)]%
        {wang2024describe}
\bibfield{author}{\bibinfo{person}{Zihao Wang}, \bibinfo{person}{Shaofei Cai}, \bibinfo{person}{Guanzhou Chen}, \bibinfo{person}{Anji Liu}, \bibinfo{person}{Xiaojian~(Shawn) Ma}, {and} \bibinfo{person}{Yitao Liang}.} \bibinfo{year}{2023}\natexlab{}.
\newblock \showarticletitle{Describe, Explain, Plan and Select: Interactive Planning with LLMs Enables Open-World Multi-Task Agents}.
\newblock   \bibinfo{volume}{36} (\bibinfo{year}{2023}), \bibinfo{pages}{34153--34189}.
\newblock
\urldef\tempurl%
\url{https://proceedings.neurips.cc/paper_files/paper/2023/file/6b8dfb8c0c12e6fafc6c256cb08a5ca7-Paper-Conference.pdf}
\showURL{%
\tempurl}


\bibitem[Wong and Hong(2007)]%
        {wong2007making}
\bibfield{author}{\bibinfo{person}{Jeffrey Wong} {and} \bibinfo{person}{Jason~I Hong}.} \bibinfo{year}{2007}\natexlab{}.
\newblock \showarticletitle{Making mashups with marmite: towards end-user programming for the web}. In \bibinfo{booktitle}{\emph{Proceedings of the SIGCHI conference on Human factors in computing systems}}. \bibinfo{pages}{1435--1444}.
\newblock
\urldef\tempurl%
\url{https://doi.org/10.1145/1240624.1240842}
\showDOI{\tempurl}


\bibitem[Zarrin and Baumeister(2015)]%
        {zarrin2015towards}
\bibfield{author}{\bibinfo{person}{Bahram Zarrin} {and} \bibinfo{person}{Hubert Baumeister}.} \bibinfo{year}{2015}\natexlab{}.
\newblock \showarticletitle{Towards separation of concerns in flow-based programming}. In \bibinfo{booktitle}{\emph{Companion Proceedings of the 14th International Conference on Modularity}}. \bibinfo{pages}{58--63}.
\newblock
\urldef\tempurl%
\url{https://doi.org/10.1145/2735386.2736752}
\showDOI{\tempurl}


\end{thebibliography}

\end{document}